%% file: acl_latex.tex
\pdfoutput=1

\documentclass[11pt]{article}

\usepackage[preprint]{acl}

\usepackage{times}
\usepackage{latexsym}

\usepackage[T1]{fontenc}

\usepackage[utf8]{inputenc}

\usepackage{microtype}

\usepackage{inconsolata}

\usepackage{graphicx}
\usepackage[utf8]{inputenc}   %
\usepackage[T1]{fontenc}
\usepackage{expex}
\usepackage{enumitem}
\usepackage{booktabs}
\usepackage{tabularx}
\usepackage{amssymb}%
\usepackage{pifont,graphicx}   %

\newcommand{\cmark}{\raisebox{0.08ex}{\scalebox{0.86}{\ding{51}}}}%
\newcommand{\xmark}{\raisebox{0.08ex}{\scalebox{0.86}{\ding{55}}}}%

\input{preamble}

\title{\THETITLE}

\author{Morris Alper\textsuperscript{*1,2} \qquad \;
  Moran Yanuka\textsuperscript{*1} \qquad \;
  Raja Giryes\textsuperscript{1} \qquad \;
  Ga\v{s}per Begu\v{s}\textsuperscript{3} \\[0.4em]
  {\footnotesize *Equal contribution} \enspace
  \textsuperscript{1}Tel Aviv University \enspace
  \textsuperscript{2}Carnegie Mellon University \enspace
  \textsuperscript{3}UC Berkeley  \enspace
   \\[0.4em]
  \url{\paperurl}
}

\begin{document}

\input{figures/teaser}

\maketitle

\input{sec/00_abs}

\input{sec/01_intro}

\input{sec/02_method}

\input{sec/03_results}

\input{sec/04_rw}

\input{sec/05_conc}

\input{sec/xx_lim}
\input{sec/xx_ack}

\bibliography{custom}

\appendix

\input{sec/xx_app}

\end{document}

%% file: preamble.tex
\usepackage{booktabs}       %
\usepackage{amsfonts}       %
\usepackage{xcolor}         %
\usepackage{lipsum}
\usepackage{amsmath}
\usepackage{cleveref}

\usepackage{graphicx}

\usepackage[utf8]{inputenc}
\usepackage[T1]{fontenc}
\usepackage{textcomp}
\usepackage{stackengine}

\usepackage{tabularx}

\usepackage{textcomp}

\usepackage{verbatim} 
\usepackage{verbatimbox}

\usepackage{listings}
\usepackage[tone]{tipa}
\usepackage[english]{babel}

\newcommand{\ourmethod}{ConlangCrafter}
\newcommand{\THETITLE}{\ourmethod{}: Constructing Languages with a Multi-Hop LLM Pipeline}

\newcommand{\paperurl}{https://conlangcrafter.github.io}

%% file: sec/00_abs.tex
\begin{abstract}

Constructed languages (\emph{conlangs}) such as Esperanto and Quenya have played diverse roles in art, philosophy, and international communication. Meanwhile, foundation models have revolutionized creative generation in text, images, and beyond. In this work, we leverage modern LLMs as computational creativity aids for end-to-end conlang creation. We introduce \emph{\ourmethod{}}, a multi-hop pipeline that decomposes language design into modular stages -- phonology, morphology, syntax, lexicon generation, and translation. At each stage, our method leverages LLMs' metalinguistic reasoning capabilities, injecting randomness to encourage diversity and leveraging self-refinement feedback to encourage consistency in the emerging language description. We construct a novel, scalable evaluation framework for this task, evaluating metrics measuring consistency and typological diversity.
Automatic and manual evaluations demonstrate \ourmethod{}'s
ability to produce coherent and varied conlangs without human linguistic expertise.
\makeatletter
\ifacl@anonymize
We will release our code and data.
\fi
\makeatother

\end{abstract}

%% file: sec/01_intro.tex
\input{figures/method}

\section{Introduction}
\begin{center}
\textcolor{black!90}{\textit{ ``p\textsuperscript{h}án dzáwali-li a-ga-galúnta-mi áta-li.''}}

{\scriptsize \textcolor{gray}{every language-{\sc intr} {\sc evid.neut-ipfv}-be\_a\_world-{\sc 3sg.intr} he/she/it-{\sc intr.}}}

\vspace{0.4em}
\textbf{``Every language is a world.''} 
{\scriptsize \textcolor{gray}{(conlang)}}
\end{center}

\smallskip

As humans have long imagined alternative methods of communication, the art of constructing languages has evolved into a creative and scholarly pursuit~\citep{schreyer2021constructed}. Constructed languages, or \emph{conlangs}, span the gamut from artistic endeavors to bring fictional worlds to life (e.g. J.R.R. Tolkien's Elvish and Dothraki in Game of Thrones) and attempts to bridge international divides for worldwide communication (e.g. Esperanto) to tests of philosophical ideas (e.g. Lojban and Toki Pona). Conlangers may spend years or even decades designing their creations, marvels of linguistic ingenuity requiring Sisyphean effort to achieve the scope and complexity of natural languages.

As foundation models are now being used for various creative tasks, including generation of novel artistic content~\citep{chakrabarty2024creativity,teleki2025survey}, we ask: Can these models be used to create conlangs?
By introducing the new paradigm of \emph{computational conlanging}, we investigate three core research questions: \textbf{(RQ1)} Can LLMs generate internally consistent linguistic systems that are distinct from those seen during training? \textbf{(RQ2)} Can LLMs generate diverse yet coherent outputs for this creative application? \textbf{(RQ3)} Can we develop objective, scalable evaluation metrics for this task with no ground-truth?

To tackle these questions, we propose a multi-hop reasoning-based LLM pipeline, \emph{\ourmethod}. This constructs a language layer by layer, using insights from linguistic typology and documentation~\citep{Genetti_2018,wals2013}, resulting in languages with diverse phonologies and grammars as shown in \Cref{fig:teaser}. Our checklist-based prompting method with injected randomness ensures diverse, typologically interesting output languages, while our proposed \emph{self-refine} loop enhances the consistency of generated language descriptions. We also propose a novel framework for \emph{constructive translation} into conlangs where lexemes and grammar may not be fully specified a priori. Finally, we introduce a novel evaluation framework that systematically assesses the quality of generated conlangs and their translations with scalable, automatic metrics. To support the validity of automatic evaluation, we assess agreement with manual expert judgements, as well as performing qualitative evaluation.

Computational conlanging offers significant theoretical and practical value. As a challenging logically-grounded task, it tests the complex meta-linguistic and logical reasoning abilities of LLMs~\cite{begus2025large} on an inherently out-of-distribution language; because the target language does not yet exist, the model cannot rely on memorized training data. From a practical perspective, it may serve as a computational creativity aid for hobbyist or professional conlangers, promising to ease laborious aspects of conlanging such as lexicon generation while allowing for creative control. Moreover, it has direct applications in procedural world or society generation in systems such as open-world video games. Finally, we see potential applications of our methodology to low-resource languages, where our framework's focus on logical consistency with set grammatical rules may be relevant for languages documented mostly through written grammars.

%% file: figures/method.tex
\begin{figure*}[t]
    \centering
    \includegraphics[width=1\textwidth]{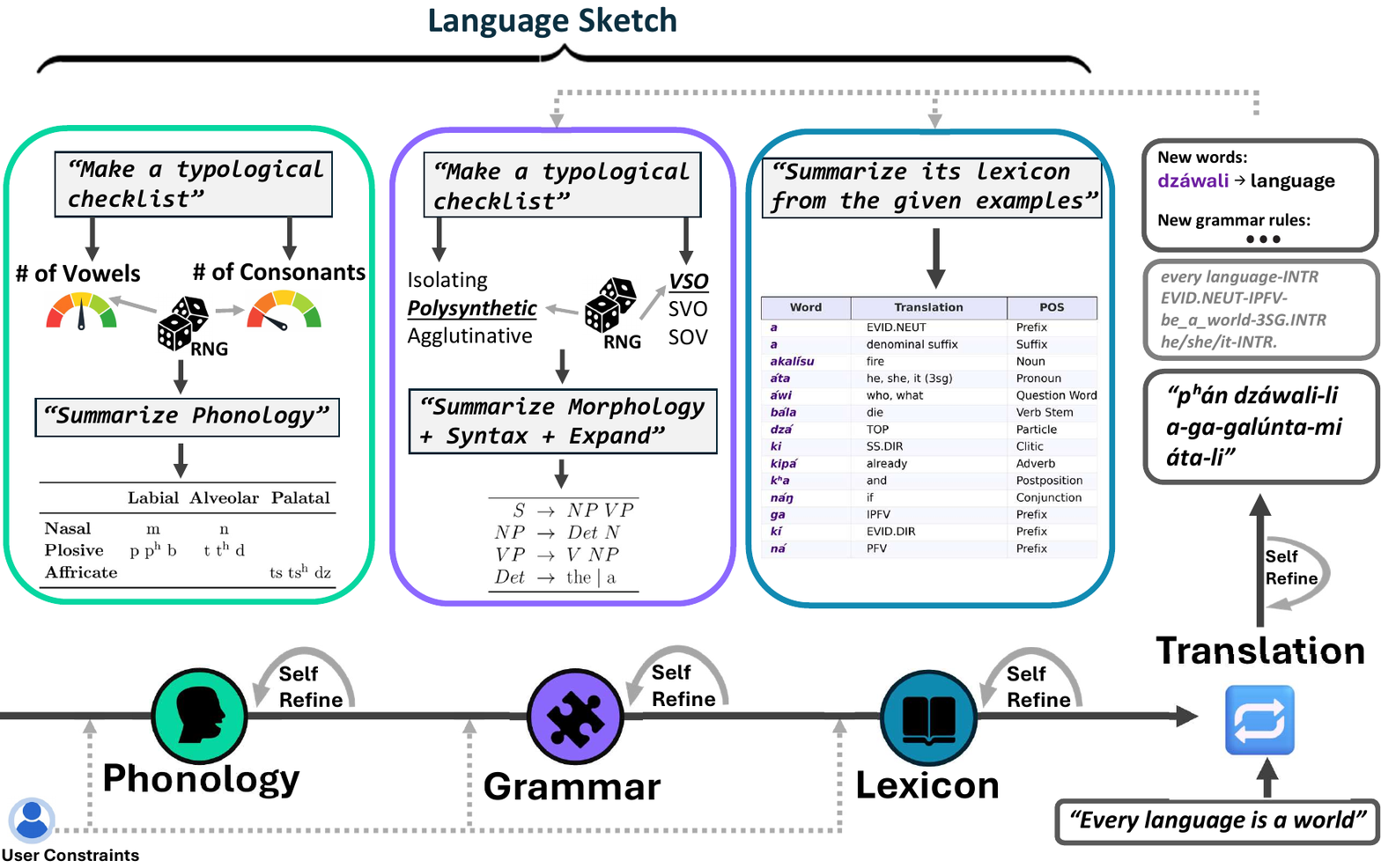}
    \vspace{-3.8em}
    \caption{
    \textbf{Our Method.}
    \ourmethod{} constructs languages with a multi-hop LLM pipeline, generating a language sketch as a sequence of linguistic layers (phonology, grammar, lexicon). We encourage typological diversity using checklist-based prompting and a random number generator (RNG), and enhance internal consistency via self-refinement. \ourmethod{} conditions on this sketch to translate and gloss new sentences, potentially including new lexical items and grammar points which may be dynamically added back to the language sketch. Prompts above are abridged and some intermediate steps are omitted.
    }
    \label{fig:method}
    \vspace{-1em}
\end{figure*}

%% file: sec/02_method.tex
\section{Method}

\ourmethod{} is an LLM-driven stochastic pipeline which generates a random conlang either fully from scratch, or in accordance with optional user-input specifications. It leverages LLMs' ability to perform creative generation that is not factually grounded, exploiting ``hallucination'' as a desirable feature, as well as exploiting their meta-linguistic knowledge~\cite{begus2025large} (i.e. ability to explicitly reason about language). As these models struggle out-of-the-box to produce diverse and coherent outputs, \ourmethod{} is designed to encourage typological diversity and to mitigate contradictions while splitting the overall task of conlang generation into tractable sub-problems. Our full method is illustrated in \Cref{fig:method}, and we proceed to describe its overall structure (\Cref{sec:pipe}) and core concepts (\Cref{sec:rand,sec:sr}).

\subsection{Overall Pipeline} \label{sec:pipe}

\ourmethod{} has the following core components

\begin{itemize}[topsep=0pt, partopsep=0pt, parsep=0pt, itemsep=0pt, leftmargin=*]
\item An LLM $M$. In practice this is a large reasoning model, i.e. a modern LLM using chain-of-thought inference-time scaling~\citep{guo2025deepseek}.
\item A memory bank $S$, referred to as the \emph{language sketch}. This consists of free text containing the language's structure description, which may be retrieved from or dynamically altered throughout the pipeline. Examples of languages sketches are provided in \Cref{sec:app_sketch}.
\item An optional user-input string $c$, with specifications or constraints on the generated conlang. By default, this is the empty string $\emptyset$.

\end{itemize}

\noindent
The pipeline proceeds in two stages:

\smallskip
\noindent
\textbf{Stage A: Language Sketch Bootstrapping.}
We first generate an initial description of the language's core structure
and store it in $S$.
Generating this description is a challenging task since it must be both sufficiently detailed and as internally consistent as possible, and our results show that a single LLM prompt is insufficient to adequately generate $S$.
To this end, we address the problem with a multi-hop pipeline that incrementally updates $S$ with linguistic information.

Linguistic theory identifies aspects of language such as phonology as objects of independent study~\citep{Genetti_2018}, and language documentation materials such as \citet{visser2022grammar} commonly split a language's description into sections based on these aspects. Following this practice, we divide a language into three key layers: \emph{phonology}, \emph{grammar} (morpho-syntax), and \emph{lexicon}. Because these layers depend on one another, we generate them sequentially (e.g., phonology precedes grammar to provide word forms). This structure parallels multi-hop approaches to other complex reasoning tasks with LLMs~\citep{khot2022decomposed}. We also note that this minimalistic structure deliberately omits additional potential layers describing linguistic aspects such as semantics, pragmatics, and orthography, which could be addressed in future work.

Each layer is produced through multiple sub-steps, prompting $M$ with the current state of $S$, optional user input $c$, and the target language aspect; the result is incorporated back into $S$. To ensure diversity and consistency, we apply randomness injection (\Cref{sec:rand}) and self-refinement (\Cref{sec:sr}). Once $S$ is initialized, it can guide translation and be expanded dynamically.

Prompts at each stage include a slot for optional user input, specifying that it takes priority
if present. This permits user control for applications such as generating full conlangs from high-level ideas or initial hand-crafted conlang sketches\footnote{Online repositories such as \url{https://conlang.fandom.com/wiki/Portal:Main} contain many partial conlang sketches, as developing complete grammars and lexicons requires substantial time and effort.}.

\smallskip
\noindent
\textbf{Stage B: Constructive Translation.}
Given $S$, \ourmethod{} translates and glosses new texts by conditioning on the explicit language description, building a corpus while dynamically updating $S$ as needed. We term this task \emph{constructive translation} as it has the unique aspect that the existing language description may be under-specified and require new, creative additions to translate a given text. This is unlike low-resource translation~\citep{tanzer2023benchmark,zhang2024hire,zhang2024can} and glossing~\citep{ginn2024can,ginn2024robustness}, in which explicit hallucination is undesirable.

During translation, \ourmethod{} prompts the model $M$ with a source text $t$ and instructions tasking it translate it consistently with the language description in $S$. Its output includes fields for the translation and interlinear gloss, as well as optional fields for new lexical items and grammar rules, which may be output as needed to resolve underspecification in $S$. The latter can be added back to $S$ to expand the language description and ensure that they are used consistently in future translations (although we without this feedback loop; see \Cref{sec:consistency_metric}). Logical consistency of these translations with $S$ is enhanced using self-refinement (\Cref{sec:sr}). This iterative process builds a translation corpus while expanding the language's grammar and lexicon.

\subsection{Randomness Injection.} \label{sec:rand}

While LLMs display
awareness of language typology, they often fail to produce diverse
outputs across runs~\citep{hopkins2023can}, resulting in limited typological variety in language sketches $S$. To address this, we inject randomness into the phonology and grammar stages during sketch bootstrapping. At the start of each stage, the LLM $M$ generates a typological checklist of ten linguistic features, each with five multiple-choice options. A random number generator (RNG) then selects one option per feature, and $M$ instantiates the language description accordingly.
This leverages the model’s metalinguistic knowledge of typology while offloading control of diversity to the external RNG.

\subsection{Self-Refinement.} \label{sec:sr}
Internal consistency is crucial, as contradictions saved to the language sketch may propagate to later stages, and translations must adhere to the language's constructed grammar. Therefore, we handle violations of logical consistency by leveraging a key observation -- evaluating generated content is often more straightforward than producing it, enabling iterative refinement through self-feedback mechanisms where LLMs model critique and revise their own outputs~\cite{wu2024meta,simonds2025self,madaan2023self}.
We adopt a similar paradigm: a critic model identifies errors and ambiguities in generated content, and an editor model revises accordingly. These are both implemented with the base LLM $M$, prompted with $S$ and the text under revision (as well as the list of identified errors in the case of the editor model). This process is repeated iteratively until no further issues are detected or a maximum number of iterations is reached to prevent infinite revision cycles.

%% file: sec/03_results.tex
\section{Experiments}

We test the utility of \ourmethod{} and its potential for creative, consistent conlang generation. We present our experimental setup (\Cref{sec:experimental_setup}); quantitative evaluation framework (\Cref{sec:quant}); results (\Cref{sec:results}); ablations (\Cref{sec:ablations}); and examples, analysis, and applications (\Cref{sec:qual}).

\subsection{Experimental Setup}
\label{sec:experimental_setup}

For base LLMs, we use DeepSeek-R1~\cite{guo2025deepseek}, and Gemini 2.5~\cite{comanici2025gemini} in two variants (Flash, Pro). For automatic evaluations, we use OpenAI o3\footnote{Described in \href{https://openai.com/index/introducing-o3-and-o4-mini/}{OpenAI's system announcement}} as the judge LLM, selected to avoid introducing bias by both generating and evaluating with the same model. For metric calculations in our experiments, we sample ${\sim}20$ languages with 10 test sentences each (see \Cref{sec:consistency_metric}); 
we confirm the sufficiency of this sample size with a statistical significance and effect size analysis (\Cref{sec:results}).
Lacking a prior method for full conlang generation, we create a reasonable baseline via a single-stage generation method. This attempts to generate a full language sketch and translations of given sentences in our format with a single prompt, without multi-hop reasoning or iterative self-refinement. Further experimental details are provided in the appendix.

\subsection{Quantitative Evaluation Framework} \label{sec:quant}

A key challenge to evaluating conlang generation is the lack of an existing framework for this novel, partially subjective task. Unlike typical machine translation, there is no inherent ground-truth for languages which do not exist. Human evaluation for attributes such as logical consistency and diversity requires expert knowledge and painstaking attention to detail, failing to scale to a sufficient sample size of languages for rigorous evaluation.

To tackle this challenge, we develop a comprehensive automatic evaluation framework using the LLM-as-a-judge approach~\cite{zheng2023judging,gu2024survey} as a scalable proxy for expert evaluation. We support its overall validity by assessing agreement with manual expert judgments performed on a smaller scale, demonstrating moderate agreement with manual evaluation on this challenging task. Despite the known limitations of such automatic metrics, this framework addresses an existing gap by enabling large-scale, quantitative evaluation of computational conlanging methods, previously infeasible with manual evaluation alone.

\subsubsection{Typological Diversity Analysis}
\label{sec:typological_diversity_metric}
To evaluate the breadth of typological variation captured by ConlangCrafter, we select a fixed set of $k=16$ basic typological features from the World Atlas of Language Structures (WALS)~\citep{wals2013} covering fundamental aspects of language. These features are chosen to cover the most fundamental and broad range of basic typological dimensions differentiating natural languages. For example, the feature ``Basic word order'' (WALS 81) indicates the relative order of subject, verb, and object in basic sentences (e.g. SVO, SOV, VSO, etc.). The full list of features used is provided in the appendix. Each feature is assigned a categorical value (e.g., ``SOV'', ``Postpositions'', ``Tonal''). While the majority of feature values can be determined, missing or underspecified values are recorded as empty for the purpose of diversity calculations.

We generate
languages $L_1, L_2, \ldots, L_N$, use a judge LLM to encode each
$L_i$ as a one-hot vector $\mathbf{x}_i \in \mathbb{Z}_2^k$ over these features, and calculate
the \textbf{diversity score} $D_{\mathrm{mean}} = \frac{2}{N(N-1)}\sum_{1\le i<j\le N}\frac{\mathrm{Ham}(\mathbf{x}_i,\mathbf{x}_j)}{k}$, where \(\mathrm{Ham}(\cdot,\cdot)\) is the Hamming distance. This reflects the average proportion of differing features between languages.

To ground these values in the distribution of existing languages, we also extract the same WALS features for all natural languages in the WALS database with sufficient coverage, yielding a sample of 1874 languages. We calculate the same diversity score on this sample and compare to values produced by \ourmethod{} and baseline methods.

\subsubsection{Internal Consistency Evaluation}
\label{sec:consistency_metric}

We evaluate internal consistency by testing for validity of a fixed set of translations conditioned on a language sketch.
This tests whether the language's rules are logically valid and can be applied coherently.
For each language, we prompt the model to translate a predefined set of $N_{t,t} = 10$ test sentences (via constructive translation, when evaluating \ourmethod{}). These are designed to cover a variety of syntactic structures (see \Cref{sec:app_sentences} for the full list of sentences). Each sentence is translated and evaluated independently, without re-adding new lexical items or grammar rules to the language sketch, to measure adherence to the original sketch. A separate judge-based evaluation prompt then assesses each of the generated translations to determine if its phonology, morphology, and syntax adhere to the rules specified in the language sketch. The \textbf{translation consistency rate}
is defined as the ratio $\frac{N_{c,t}}{N_{t,t}}$ of correctly formed translations to total translations, averaged over all generated languages.
While this is limited in equally penalizing minor and major errors, it nevertheless
provides an indication of the degree to which the language can be parsed.

\input{tables/results_automatic}

\input{figures/tsne_typological_diversity}

\subsubsection{Manual Expert Evaluation}
\label{sec:manual}

Human evaluation in our setting is only feasible on a small scale, as it requires cross-referencing with a language sketch in order to assess language traits and consistency of a translation with all relevant details. Unlike standard NLP annotation tasks, each judgment requires a holistic consistency check against a novel, custom-designed grammar, rather than comparison to a fixed reference. In addition, this may only be performed by experts with linguistic training to understand technical linguistic materials. To support the validity of our larger-scale automatic evaluation, we perform a manual human evaluation parallel to \Cref{sec:typological_diversity_metric,sec:consistency_metric} on all $N=20$ languages and $N_{t,t}=10$ translations output by a single model. We briefly describe the annotation procedure below, with further details and full task instructions provided in the appendix.

For evaluating diversity calculations, annotators received browsable HTML files containing each language's sketch and automatically inferred WALS features. They were instructed to label each inferred feature based on correctness given the language sketch. We report the error rate of automatic WALS features relative to these annotations as well as inter-annotator agreement (Cohen's $\kappa$).

Translation consistency was evaluated similarly, with browsable HTML files containing language sketches and candidate translations paired with interlinear glosses. An automated LLM judgement was displayed alongside each candidate as an auxiliary cue only; the annotation task instructions requested the annotators to rely on their own reading of the language sketch when scoring. While we note that this may potentially introduce a priming effect, our pilot tests found that performing this manual evaluation entirely from scratch was infeasibly labor-intensive. We adopt this design as a practical compromise, as well as noting that annotators readily corrected automatic judgements as reflected in agreement metrics. We report agreement (Spearman $\rho$) between mean manual and automatic scores. These are both on ordinal scales from 0=\emph{fully inconsistent} to 2=\emph{fully consistent} (noting that the final automatic metric in \Cref{sec:consistency_metric} only counts fully consistent translations). We also report inter-annotator agreement (weighted $\kappa$).

Our manual evaluation was performed by two PhD students in linguistics with expertise in language documentation and typology, recruited through our institution. Participant compensation complied with applicable institutional policies. Each annotation required detailed expert analysis, involving cross-referencing between the generated sentence, its gloss, and the full language sketch. In total, the manual evaluation required approximately 35 hours of expert labor by PhD-level linguists. The goal of manual evaluation is therefore not to provide large-scale human scoring, but to validate that our automatic metrics meaningfully track expert judgments on this challenging task.

\input{tables/component_ablations}

\input{tables/temperature_ablation}

\input{tables/qual}

\subsection{Results}
\label{sec:results}

\paragraph{Quantitative Results.}
Quantitative results are shown in \Cref{tab:results_automatic}, evaluating \ourmethod{} with base LLMs as well as the baseline method (\Cref{sec:experimental_setup}). Our method outperforms the baseline in both diversity and consistency metrics. 
For diversity scores, all comparisons between the baseline and our method are highly significant (two-sided t-tests, $p < 10^{-6}$) with large effect sizes (Cohen's $d > 3.0$). While features may be underspecified, we note mean feature coverage of 94--98\% across settings, indicating a negligible impact on diversity scores. For consistency scores, baseline comparisons are significant (two sided t-tests, paired by sentence) for Gemini-2.5-Pro ($p \approx 0.023$) and DeepSeek-R1 ($p \approx 0.014$) with large effect sizes ($d > 1.4$). For Gemini-2.5-Flash, the small effect ($d \approx 0.37$) does not reach significance ($p \approx 0.294$) at our sample size, but is consistent with findings that \ourmethod{} increases diversity without reducing consistency.
In addition, we see moderate differences between the different LLMs in performance, with the stronger Gemini-2.5-Pro and DeepSeek-R1 performing better at this challenging task. Overall, generation of fully consistent translations remains a challenge, although our full method yields a better diversity-consistency tradeoff relative to the baseline.

In Figure \ref{fig:tsne}, we apply t-distributed stochastic neighbor embedding (t-SNE) to visualize the pairwise Hamming distance matrix in two dimensions, where spatial proximity reflects typological similarity. The scattered distribution of languages across the 2D space visually confirms the high typological diversity compared with the baseline. These findings confirm that \ourmethod{}'s randomness injection and modular prompting yield a diverse typological sample, even when evaluated against a fixed, expert‐selected set of WALS features.

\paragraph{Comparison to Natural Languages.}
We also calculate diversity values for natural languages (\Cref{sec:typological_diversity_metric}), yielding $D = 0.43 \pm 0.003$, which falls between baseline and \ourmethod{} values in \Cref{tab:results_automatic}. All pairwise comparisons are statistically significant (t-tests, all $p < 10^{-6}$). This indicates that the baseline method generates languages with lower typological diversity than natural languages, while \ourmethod{} exceeds this level of diversity.
Thus, despite our base LLMs exhibiting strong performance on general multilingual benchmarks, they fail to match the diversity of natural languages when used without our method, failing to leverage their knowledge of cross-lingual variation.
By contrast, our method
demonstrates successful exploration of the full typological space including novel feature combinations. For creative generation applications, exceeding natural language diversity is desirable, as it allows for imaginative world-building and testing of linguistic hypotheses with typologically unusual combinations.

To further assess novelty of generated conlangs relative to natural languages, we select the nearest neighbor to each generation in the WALS database of natural languages via Hamming distance. For \ourmethod{} with Gemini-2.5-Pro, these nearest neighbors have a mean of 56.4\% matching comparable features (min=50.0\%, max=68.8\%), showing that the conlangs exhibit highly novel typological feature combinations rather than replicating existing languages. For example, the four languages in \Cref{tab:qualitative} are closest to Yaqui (Uto-Aztecan, 52.9\% match), Canela (Macro-Ge, 56.2\%), Hausa (Chadic, 56.2\%), and Axininca (Arawakan, 53.8\%), respectively. These values reflect qualitative divergence in basic typological features; for example, the first language shares Yaqui's simple tone system, adjective-noun order, and question particle strategy, while differing in basic word order, adposition type, and consonant inventory size.

\paragraph{Manual Evaluation Results.}
For manual evaluation of diversity calculations, we examined automatic WALS features for all cases where annotators agreed (Cohen's $\kappa=0.58$; moderate agreement). The most common issue was due to underspecification issues; 12\% of potential features were mistakenly left unspecified, while 4\% of features that were not specified had incorrectly inferred automatic values. Excluding underspecification issues, automatic feature extraction achieved 91\% accuracy. This demonstrates that our diversity metric is based on largely accurate typological analysis, with the main existing limitation regarding the treatment of underspecified features.

For manual evaluation of translation consistency, we find significant agreement between manual and automatic metrics, with Spearman $\rho = 0.68$ ($p < 10^{-27}$). Inter-annotator agreement is quadratic weighted $\kappa = 0.43$ ($p < 10^{-4}$), indicating moderate agreement between annotators. This level of agreement is expected, as judgments require interpreting underspecified grammars and weighing minor versus major inconsistencies, decisions for which multiple valid analyses may exist. Overall, automatic scores tend to be stricter than manual evaluations, but still reasonably track human judgements when calibrated for this strictness level.

We contextualize both of these moderate agreement scores by noting that similar scores are standard in evaluations of natural language generation~\citep{van2021human}, particularly for creative tasks which involve subjective reasoning and natural variability of human linguistic interpretation. In this context, exceptionally high agreement may be a negative indicator, signaling overly simplistic evaluation criteria~\citep{amidei2018rethinking}.

\subsection{Ablation Analysis}
\label{sec:ablations}
In \Cref{tab:ablation_gemini_pro} we ablate
key components of our system: multi-hop reasoning (MH), randomness injection (RNG),
and iterative self-refinement (SR). We add each of these in turn to the baseline method (described in \Cref{sec:experimental_setup}) and observe their cumulative effect.
Multi-hop reasoning and randomness injection significantly improve diversity, key for generating typologically interesting and varied languages. This reduces consistency, as expected, since diverse outputs may be more logically challenging. However,
self-refinement
mitigates this issue, significantly increasing consistency approaching or beyond baseline levels. This matches qualitative observations that multi-hop reasoning and randomness injection are key for interesting, diverse outputs, while self-refinement succeeds in correcting many salient logical errors in generations.

To further test the importance of
randomness injection,
we compare to
standard diversity-enhancing techniques reliant on sampling temperature, under our generation settings (without self-refinement). While increasing the sampling temperature (using nucleus sampling with $p=0.95$) effectively increases variation, it operates at the token level rather than the typological level. As shown in Table~\ref{tab:ablation_creativity}, approaching the diversity score of our method ($0.60$) via standard sampling requires raising the temperature to $T \approx 1.4$ resulting in a sharp trade-off with consistency, which crashes to $0.34$.
By contrast, our method injects randomness directly into the typological feature selection (e.g., randomly sampling relative order of subject, object, and verb), allowing the language model to generate the description at a lower temperature ($T=0.6$) to ensure logically coherent and fluent text. This maintains higher internal consistency ($0.43$) while achieving high typological diversity ($0.60$).
We further note that alternative diversity-enhancing methods, such as specialized decoding strategies \cite{hewitt-etal-2022-truncation, chang2025real} or internal interventions \cite{chung2025modifying, zhou-etal-2025-bridging}, typically require access to model logits or gradients. These are not feasible in our setting as we focus on state-of-the-art large reasoning models, which are either closed-weights (e.g., Gemini-2.5-pro) or prohibitively large to run locally (e.g. DeepSeek-R1).

\subsection{Qualitative Analysis and Application}
\label{sec:qual}

Beyond metrics,
we examine the creative and linguistic quality of
generations.
Table~\ref{tab:qualitative} illustrates unconditional \ourmethod{} generations.
The incorporation of unique typological features demonstrates the system's grasp of cross-linguistic variation beyond major world languages, and its effectiveness at diversity between generations.
Table~\ref{tab:qualitative} also includes two examples of languages generated with user-input constraints, demonstrating user control over the generation process. These include an all-vowel language in which there are no consonant phonemes, and a language designed for an alien cephalopod species in which sound-based phonology is entirely replaced by color-based ``chromemes'' and gesture-based ``kinemes.'' These examples illustrate that \ourmethod{} can accommodate creative and speculative constraints far beyond the typological space of natural languages, enabling applications in world-building and theoretical linguistic exploration. Additional controlled examples are provided in the appendix (\Cref{tab:qual_controlled}).
Along with these promising qualitative  results, we note that quantitative evaluation of faithfulness to user-input constraints remains an open challenge, as they are inherently open-ended, variable in scope, and lack a definitive ground truth for automatic evaluation.

\makeatletter
\ifacl@finalcopy
We provide a HTML interface to view these and other generated languages on our project page.
\fi
\makeatother

%% file: tables/results_automatic.tex
\begin{table*}[t]
  \centering
  \setlength{\tabcolsep}{3pt}
  \begin{tabularx}{\linewidth}{@{}l
    *{8}{>{\centering\arraybackslash}X}@{}}
    \toprule
    &
    \multicolumn{2}{c}{\textbf{Gemini-2.5-Flash}}
    & \multicolumn{2}{c}{\textbf{Gemini-2.5-Pro}}
    & \multicolumn{2}{c}{\textbf{DeepSeek-R1}} \\
    \cmidrule(lr){2-3}\cmidrule(lr){4-5}\cmidrule(lr){6-7}\cmidrule(lr){8-9}
    \textbf{Metric}
      &
      \textbf{Baseline}
      & \textbf{Ours}
      & \textbf{Baseline}
      & \textbf{Ours}
      & \textbf{Baseline}
      & \textbf{Ours} \\
    \midrule
    Diversity ($\uparrow$)
      & $0.25 \pm 0.01$
      & $0.60 \pm 0.01$
      & $0.26 \pm 0.02$
      & $0.58 \pm 0.01$
      & $0.35 \pm 0.01$
      & $ 0.56 \pm 0.02 $ \\
    Consistency ($\uparrow$)
      & $0.30 \pm 0.04$
      & $0.38 \pm 0.04$
      & $0.32 \pm 0.05$
      & $0.54 \pm 0.05$
      & $0.21 \pm 0.04$
      & $0.39 \pm 0.05$ \\
    \bottomrule
  \end{tabularx}
  \caption{\textbf{Typological diversity
  and
  consistency} comparison between the baseline method and ConlangCrafter,
  measured using the automatic metrics described in Section \ref{sec:quant}. Results are shown with $\pm$ stderr.
  }
\label{tab:results_automatic}
\end{table*}

%% file: figures/tsne_typological_diversity.tex
    \begin{figure}[t]
        \centering
        \includegraphics[width=\columnwidth]{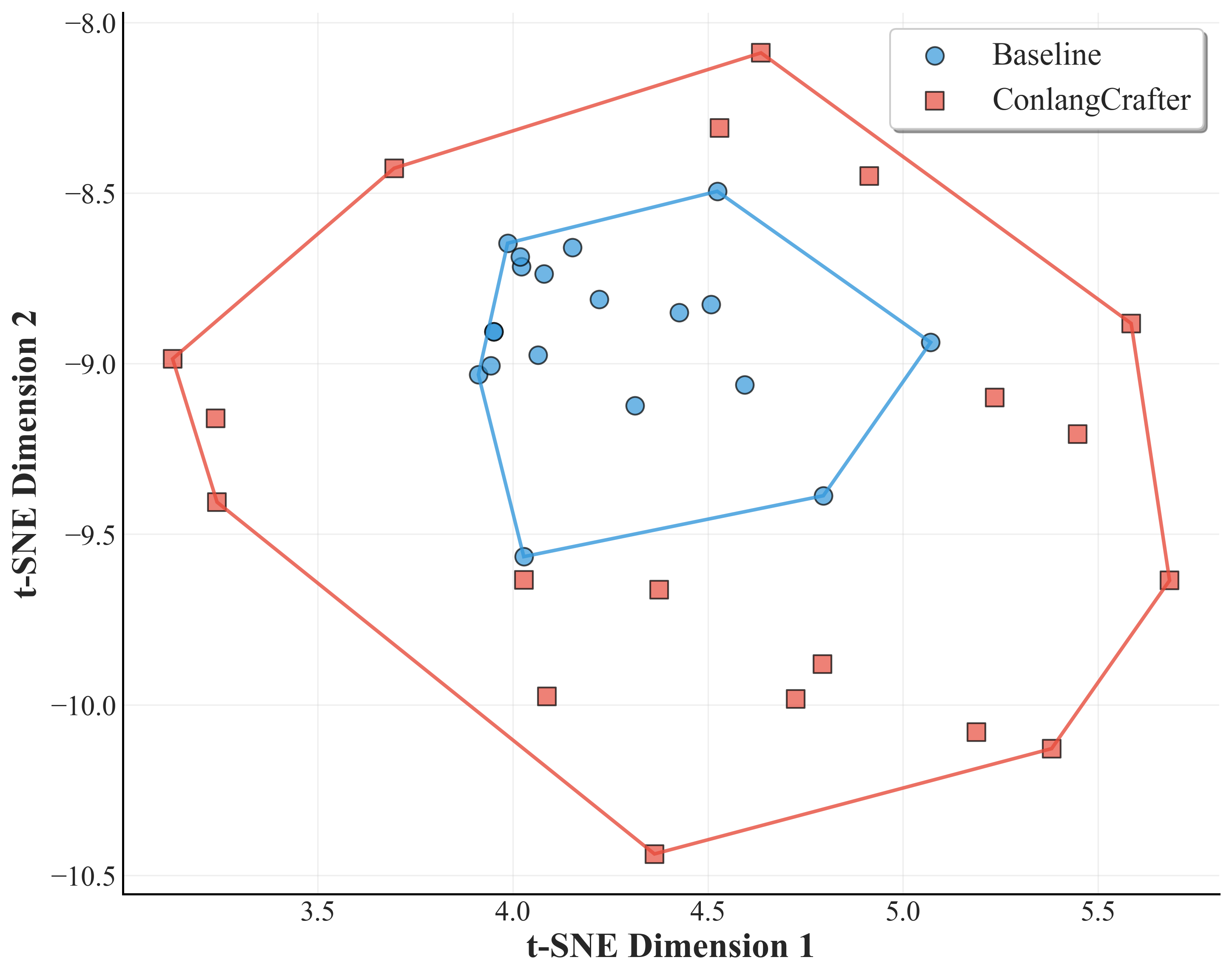}
       \caption{
        \textbf{t-SNE visualization of typological diversity.} 
       Each point represents a generated language, where spatial proximity reflects typological similarity based on WALS features.
       The dispersed distribution of \ourmethod{}-generated languages highlights their higher typological diversity.
       }
        \label{fig:tsne}
    \end{figure}

%% file: tables/component_ablations.tex
\begin{table*}[t]
  \centering
  \begin{tabularx}{\linewidth}{@{}l
    *{4}{>{\centering\arraybackslash}X@{}}}
    \toprule
    \textbf{Metric}
      & \textbf{Baseline}
      & \textbf{+MH}
      & \textbf{+MH+RNG}
      & \textbf{+MH+RNG+SR (Full)} \\
    \midrule
    Diversity ($\uparrow$)
      & $0.26\pm 0.02$
      &  $0.52 \pm 0.01$
      &  $0.60 \pm 0.02$
      &  $0.58 \pm 0.01$\\
    Consistency ($\uparrow$)
      & $0.32 \pm 0.05$
      & $0.40 \pm 0.05$
      & $0.43 \pm 0.05$
      & $0.54 \pm 0.05$\\
    \bottomrule
  \end{tabularx}
  \caption{\textbf{Ablation study of \ourmethod{}'s components} showing the cumulative effects of adding key components -- Multi-Hop reasoning (MH), RNG randomness injection, and Self-Refinement (SR) -- on automatic metrics over 20 languages generated by Gemini-2.5-Pro. Results are shown with $\pm$ stderr.}

\label{tab:ablation_gemini_pro}
\end{table*}

%% file: tables/temperature_ablation.tex
\begin{table}[t]
  \centering
  \setlength{\tabcolsep}{1.5pt}
  \begin{tabularx}{\linewidth}{@{}l*{3}{>{\centering\arraybackslash}X}@{}}
    \toprule
    \textbf{Temp.} & \textbf{RNG} & \textbf{Diversity} & \textbf{Consistency} \\
    \midrule
    0.6 & \xmark  & $0.46 \pm 0.02$ & $0.59 \pm 0.08$ \\
    1.0   & \xmark  & $0.50 \pm 0.02$ & $0.40 \pm 0.09$ \\
    1.4 & \xmark  & $0.56 \pm 0.02$ & $0.34 \pm 0.03$ \\
    1.6 & \xmark  & $0.54 \pm 0.02$ & $0.25 \pm 0.06$ \\
    \midrule
    0.6 & \cmark & $0.60 \pm 0.02$ & $0.43 \pm 0.05$ \\
    \bottomrule
  \end{tabularx}
\caption{\textbf{Comparison of diversity enhancement methods:} temperature sampling (above) and our RNG-based randomness injection (below).
We
achieve high diversity without the severe consistency penalty associated with high-temperature sampling. Self-refinement was disabled in this ablation.} 

  \label{tab:ablation_creativity}
\end{table}

%% file: tables/qual.tex
\begin{table*}[t]
\centering
\small
\begin{tabularx}{\linewidth}{@{}>{\raggedright\arraybackslash}X>{\raggedright\arraybackslash}X>{\raggedright\arraybackslash}X@{}}
\toprule
 \multicolumn{2}{c}{\textbf{Sample Sentences}}
 &
 \textbf{Key Features} \\
\textbf{``The big dog is sleeping.''} & \textbf{``She will give him water.''} & \\
\midrule

\textipa{tsO\tone{1} gO-snat\tone{5} k\:Ra\tone{515} !a\tone{51}.TO\tone{1}}
\newline {\newline PRS 3SG.NHUM.S-sleep big dog.NOM}

&

\textipa{Si\tone{1}.\|[tu\tone{5}-i\tone{51} go\tone{1}-|o\tone{15} do\tone{1}-sul\tone{1}-bo\tone{5}-dZo\tone{5}}
\newline {man-ACC 3SG.HUM.S-FUT 3SG.HUM.O-water-have-CAUS}

& 
Click consonants, polysynthetic morphology, noun incorporation, OVS word order, switch-reference, double marking

\\
\midrule

\textipa{nO-k'a\super{n}da jOk'O to-hoto \super{m}bulo hE}
\newline PAT-sleep dog.SG.ABS 3SG.N.ABS-be.STAT big.ADJ PRF

&

\textipa{se tU-nO-p'ajE hE wuwolo hE tu}
\newline IRR AGT-PAT-give 3SG water 3SG to

&
ATR harmony, ejectives and prenasalized stops, active-stative alignment, VSO word order, ejective-conditioned word order changes
\\
\midrule
\textipa{"g\super{w}aNk "k\super{j}esp@ n\super{j}o.ska"m\super{j}i Si}
\newline big dog sleep 3S.PAT

&

\textipa{n\super{w}a"co.l\super{j}o "gr\super{j}o.b@ k\super{w}o Si Si}
\newline FUT give 3S.AGT 3S.PAT 3S.PAT

&

Frequent secondary articulations, isolating morphology, evidentiality marking, SOV word order with post-verbal particles marking person
\\

\midrule
\textipa{\:s\v{a}N \|[t\'{@}m-j\'{@} \:n\^{a}.\:t\`{a}-s\v{a}-n}
\newline dog big-ERG.ANIM sleep-IPFV.VIS.LOW-T-3SG

&

\textipa{n\'{@}.k\`{a} q\'{a} m\v{a}.\|[n\v{a}.a l\'{@}m.@t kj\^{a}.l\`{a}.n}
\newline FUT to man-ABS.ANIM water-ABS.INAN give-PFV.VIS-3SG
&
Vertical vowel system, dental-alveolar-retroflex distinction, split ergativity by aspect, tonal polarity in verb inflection, animacy-governed adposition order, serial verb constructions
 \\

\midrule
\multicolumn{3}{l}{\textit{User-controlled generation (user input in italics):}} \\
\midrule

\textit{``There are no consonant phonemes.''}

&

\textipa{a-A.o.-A-\'{u}\textsubarch{u}-\'{o}\textsubarch{u}}
\newline 1.SUBJ-see-2.OBJ-PFV-DIRECT
\newline ``I have seen you.''

&

Non-phonemic glottal and pharyngeal onsets appear phonetically depending on syllable nucleus vowel quality.
\\

\midrule

\textit{``The language is produced by an alien cephalopod species. Phonemes are color values and gestures rather than consonants or vowels.''}

&

PB BR SRHC.PWHC(Peak) SRSW(Peak)
\newline PFV VIS CL1.AGT-move(V.PEAK) this.one(CL1)
\newline ``Did this being move?''

&

``Chromemes'' (color-based) have codes indicating dynamic features like static (S-) and pulsating (P-), and hue features like warm (-R) and cool (-B). ``Kinemes'' (gesture-based) such as HC involve curling tentacles. Contour ``tones'' indicate movement patterns through the water.
\\

\bottomrule
\end{tabularx}
\caption{\textbf{Qualitative examples} of ConlangCrafter-generated languages. The first four were generated unconditionally; the last two use user-input constraints (shown in italics), demonstrating user control and the ability to test creative ideas beyond attested natural languages.}
\label{tab:qualitative}
\vspace{-1em}
\end{table*}

%% file: sec/04_rw.tex
\section{Related Work}

\noindent
\textbf{Computational Conlanging.}
While several software tools\footnote{Such as those listed at the \href{https://www.frathwiki.com/Software_tools_for_conlanging}{FrathWiki}} have been created by the conlanging community to automate various aspects of conlanging, 
machine learning-based methods are limited and piecemeal. 
Prior works have explored generating words in isolation~\citep{zacharias2022extending}, assigning or interpreting visual associations with nonsense words~\citep{alper2023kiki,matsuhira2024investigating,kouwenhoven2025shaping}, and the ability of neural models to learn constructed languages with controlled linguistic properties~\cite{mccoy2018revisiting,mccoy2021infinite,kallini2024mission,mccoy2025modeling} or constructed via cryptographic substitutions~\cite{marmonier2025explicit}. Recent work has also studied statistical properties of generated text when prompting LLMs to construct languages~\citep{marmonier2025explicit}, but this is done in a single prompt and does not output fine-grained interpretable linguistic structure.
By contrast, our approach constructs languages end-to-end with LLMs, with linguistically interpretable structure enabling downstream translation.

\smallskip
\noindent 
\textbf{Computational Creativity.}
Generative models are increasingly being leveraged for creative tasks, with recent works exploring
the use of LLMs for applications such as story generation~\citep{venkatraman2025collabstory,teleki2025survey}, collaborative narrative modeling~\citep{qiu2025deep}, and research ideation~\citep{si2025can}.
Besides fully autonomous systems, human-computer interaction studies have explored the use of generative models as a creativity aid~\citep{chakrabarty2024creativity,kumar2025human}. Similarly to our work, some studies have explored the invention of novel words~\citep{malkin2021gpt} and visual concepts~\citep{richardson2024conceptlab}. A key challenge is evaluating performance on tasks which are inherently open-ended, which has motivated
various
evaluation frameworks and benchmarks for creativity~\citep{li-etal-2025-automated-creativity,hou2025creativityprism,zhao2025assessing}.
In this vein, we address
a novel creative domain that requires both typological diversity and strict logical consistency, distinct from typical creative generation tasks.

\smallskip
\noindent
\textbf{Diverse Text Generation.} 
LLMs often suffer from limited diversity, as common decoding strategies fail to match human text statistics~\citep{holtzmancurious2020},
potentially exacerbated by post-training~\citep{kirk2023understanding,yun2025price}.
Mitigation strategies
include
temperature, nucleus~\citep{holtzmancurious2020}, and min-p sampling~\citep{nguyen2024turning}, as well as handcrafted prompting techniques~\citep{tian2024macgyver,wang-etal-2025-multilingual,hu2025dipper}. 
These address token-level or semantic diversity, while we require targeted structural diversity at the level of linguistic typology.%

%% file: sec/05_conc.tex
\section{Conclusion}

We have proposed the novel paradigm of computational conlanging, showing that \ourmethod{} can construct coherent artificial languages through a novel multi-hop pipeline, validated by a new scalable evaluation framework for this task as well as manual expert judgements. \ourmethod{} offers a new computational creativity tool for language construction, enabling user-guided conlang creation with potential future applications such as procedural society generation and distillation of meta-linguistic reasoning for low-resource language NLP. We foresee extensions such as scaling to larger grammars and lexicons, exploring additional language aspects (e.g. semantics, multi-modality), extending the set of models tested to create a comprehensive benchmark, and modeling languages as evolving communication tools between agents across time, space, and culture.

%% file: sec/xx_lim.tex
\section*{Limitations}

While our pipeline is designed to encourage diversity and avoid collapse to existing languages, LLMs may still be biased towards English and other high-resource languages
in ways not captured by our evaluation framework, paralleling known issues with LLMs applied in multilingual settings~\citep{chen2024good,singh2025global}.
The limited nature of linguistic typological information seen during training (and the limited coverage of the worlds' languages in the linguistic literature overall) may prevent our method from generating extremely unusual features. Moreover, LLMs cannot capture the full range of human creativity and may lack the insight to invent novel, higher-level philosophical ideas.

Our language summaries only describe fundamental language components (phonology, morpho-syntax, lexicon) while disregarding aspects such as semantics, pragmatics, discourse strategies, and orthography, which could be added in future work.
These summaries are too short to capture a fraction of the complexity occurring in real languages; future research could study how to scale up their size, which currently is incompatible with our prompting methods which are limited by the context length of current LLMs.

Our evaluation method is limited
due to the extensive labor required for manual evaluation, the potential for annotator priming from displayed automatic judgments (see \Cref{sec:manual}), the open challenge of quantifying adherence to user-input constraints (\Cref{sec:qual}), and the overall difficulty in formulating automated metrics for conlanging. Future work could design more scalable evaluation frameworks for this task.
Additionally, our pipeline incurs a substantial computational cost due to repeated LLM calls with long contexts, particularly in self-refinement loops; future work could distill \ourmethod{} to achieve similar results with efficient inference.

\section*{Ethical Considerations}

As with other uses of LLMs and other generative models, our method requires responsible use and disclaimers accompanying generated content to avoid disseminating misinformation. This includes concerns such as potential use of generated conlangs to evade content moderation, and possible cultural misrepresentation when generating languages inspired by real-world cultures. Repeated calls to LLM generation may consume significant compute resources, and we look to future work to improve the efficiency of our method. Finally, while our work may have future applications to low-resource languages, we emphasize that generation of fictional content must not come at the expense of research attention and resources being directed toward living communities.

Regarding manual annotation, no personally identifying information or sensitive data were collected, and only aggregated results are reported. We did not obtain IRB approval because the research does not involve human subjects. Individuals' participation was limited to professional linguistic annotation, and the research questions concern language model behavior, not the annotators.

%% file: sec/xx_ack.tex
\section*{Acknowledgments}

We thank Alexander Elias, Allegra Robertson Molinaro, Kai Schenk, and Wesley Kuhron Jones for their linguistic assistance. We also thank Eric Chen and Hanzhi Zhu for their helpful feedback.

%% file: sec/xx_app.tex
\input{tables/wals_features}

\section{Implementation Details}

\subsection{Pipeline Design and Prompts}

Full prompts used in our pipeline
are provided in \Crefrange{fig:phonology_prompts}{fig:translation_prompt}.
In each prompt, fields in curly braces \{ \} are values that are filled in at inference time; these include configuration variable (e.g. \texttt{\{n\_questions\}} has default value 10) as well as outputs of previous steps (e.g. \texttt{\{checklist\}} in the second prompt in \Cref{fig:phonology_prompts} is filled in with the output from the previous prompt).

Unless otherwise specified, the prompts in 
\Crefrange{fig:phonology_prompts}{fig:translation_prompt} are run serially. The last lexicon prompt (bottom of \Cref{fig:lexicon_prompts}) is run in a loop until the lexicon contains a minimum number of items (100 by default). The self-refinement prompts (\Crefrange{fig:sr_prompts}{fig:sr_translation_prompts}) are run after corresponding steps; the critic prompts there are used to identify issues, and the amendment prompts edit to address these issues. These are run in a loop until the critic returns an overall score above a fixed threshold.

We proceed to describe core design choices motivating this pipeline, beyond the overall multi-hop structure, randomness injection, and self-refinement components described in the main paper. \Cref{fig:phonology_prompts} includes a third step discussing word shapes to encourage diverse word shapes as we find that otherwise languages are prone to collapsing to monosyllables (typologically attested but limiting diversity of generations). \Cref{fig:grammar_prompts} includes an expansion step to achieve a more satisfying level of detail in the output grammar, with an additional merger step to holistically incorporate these details into the grammar. Many prompts explicitly request interesting and unusual outputs, which complements our goal of constructing diverse and interesting languages.

\subsection{Baseline Prompt}

The single prompt used for baseline comparisons is given in \Cref{fig:baseline_prompt}. This uses the same output fields as the full \ourmethod{} pipeline, while requesting an entire language sketch and all sentence  translations in a single prompt.

\subsection{Evaluation Prompts}

The prompts used for evaluation metrics are shown in \Crefrange{fig:diversity_prompt}{fig:consistency_prompt}. This includes the prompt used for WALS feature extraction and the prompt used for the judge evaluating translation consistency.

\subsection{Constructive Translation}
\label{sec:app_sentences}

For constructive translation, we translate the following fixed sentences which include a variety of syntactic structures as detailed below:
\begin{itemize}
\item The big dog is sleeping. \textit{(Attributive adjective, present continuous, definite article)}
\item Where is my book? \textit{(WH-interrogative, possessive, copula)}
\item She will give him water. \textit{(Ditransitive, future tense, pronouns)}
\item This child walked to that house yesterday. \textit{(Demonstratives, past tense, directional, temporal adverb)}
\item Are you hungry? \textit{(Yes/no question, adjectival predicate)}
\item Give me the red bird! \textit{(Imperative, indirect object pronoun, attributive adjective)}
\item The woman and the man are talking. \textit{(Coordination, present continuous)}
\item I do not see three cats. \textit{(Negation, transitive, numeral, plural)}
\item There is a black mountain. \textit{(Existential construction, indefinite article, attributive adjective)}
\item Two children played in the garden. \textit{(Numeral, past tense, locative prepositional phrase)}
\end{itemize}

New lexical items and grammar may be appended to the language sketch, though in our evaluations we translate each sentence independently (without re-adding new lexical items and grammar rules to the language sketch).

\subsection{WALS Features and Visualization}

WALS features that are extracted and used for typological diversity calculations are shown in \Cref{tab:linguistic_features}. We include their numbers from the WALS database, along with an overall description of the contents of each feature. The visualization in \Cref{fig:tsne} uses default scikit-learn t-SNE hyperparameters.

\subsection{Language Generation Details}

LLM generation uses the maximum values for maximum output tokens and otherwise using default recommended hyperparameters for decoding (temperature, top-p). For metric calculations in our experiments, we sample $N \approx 20$ languages.

Due to resource constraints, the number of languages generated for metric calculations varies depending on setting ($16$ for Gemini-2.5-Flash, $20$ for Gemini-2.5-Pro, and 24 for DeepSeek-R1). We use rejection sampling to remove degenerate outputs that do not conform to the required language sketch format, rejecting about 5-10\% of the samples, depending on the model.

For self-refinement, the critic provides an overall score on a 1--10 scale, with 9 being explicitly given as the score for content which is fully consistent but may contain minor unclear points. Self-refinement terminates when
this score exceeds a fixed threshold (9 for language sketch generation, 10 for translation),
or after 10 iterations.

\input{tables/qual_controlled}

\section{Additional Results}

\subsection{Example Language Sketches}
\label{sec:app_sketch}

\makeatletter
\ifacl@anonymize
In the supplementary material,
\else
On our project page,
\fi
\makeatother
we provide various \ourmethod{}-generated language sketches and translations as a browsable HTML page.
In addition, we provide the full text of a single language sketch and constructive translation outputs in \Crefrange{fig:sample_sketch_p1}{fig:sample_sketch_p7}. Note that the sketch is saved as free text. In practice, the phonology and grammar sections are markdown-formatted, while the lexicon is in CSV format.

\subsection{Qualitative Results with User-Input Constraints}
In \Cref{tab:qual_controlled}, we show additional qualitative examples from languages generated with user-input constraints (beyond those presented in the main paper in \Cref{tab:qualitative}), which our system accepts as free text. These illustrate that our system succeeds in generating creative conlangs that adhere to the user-input specifications, allowing users to test creative,  linguistic, and philosophical ideas beyond what is attested in natural languages.

\section{Computation Usage}

As our tests focus on large models run via external APIs, we use token counts as the primary measure of computation. This value varies depending on the model used and due to randomness across runs. When using DeepSeek-R1, our full pipeline, including self-refinement and sentence translations, consumes approximately 660K tokens for a single language, costing about $4$ USD on the Together AI platform. This cost is largely driven by self-refinement loops; generating a single language sketch without self-refinement or sentence translations consumes approximately 70K tokens, representing a ten-fold decrease. By comparison, using Gemini 2.5 Flash, generating a complete language with our standard translation set requires an average of approximately 2.15M tokens. In this case, most of the cost is incurred during the translation stage and within the iterative self-refinement QA loops, which are critical for maintaining internal consistency.

\section{Annotator Instructions and Protocol}
\label{app:annotation-protocol}

\subsection{Manual Diversity Evaluation}

\noindent
\textbf{Materials shown to annotators.}
Annotators were shown language sketches (phonology and grammar sections only) along with automatically-inferred WALS features. Each feature was shown with the chosen value highlighted, as well as all possible values for that feature.

\noindent
\textbf{Full Instructions Provided.}

\emph{The browsable HTML file includes 20 generated languages. Each is shown with its description (phonology and grammar) followed by automatically-inferred typological (WALS) features in the section "WALS Features for Diversity Calculation". Your task is to review these automatic features and validate whether they are correctly inferred from the language description.}

\emph{In the spreadsheet, for each language and feature, please indicate one of the following values:}
\begin{itemize}
\item \emph{0: If the automatic feature value is a mistake (e.g. automatic value said SOV but language is actually SVO)}
\item \emph{1: If the automatic feature value is correct}
\item \emph{2: If the automatic feature value is undefined but should be defined (e.g. automatic value is empty but should be SVO)}
\item \emph{3: If the automatic feature value is defined but should be undefined (e.g. automatic value is SVO but word order is actually not defined anywhere in the language sketch)}
\end{itemize}

\subsection{Manual Consistency Evaluation}

\noindent
\textbf{Materials shown to annotators.}
Annotators were shown full language sketches along with constructive translation results. Each translation entry presented:
\begin{enumerate}
  \item the source English sentence (prompt),
  \item the generated target sentence (surface form),
  \item an interlinear gloss,
  \item optional fields documenting any \emph{new lexical items} or \emph{new grammatical rules} the generator introduced,
  \item an automatically produced LLM judgement (inconsistent / mostly consistent / consistent) with a brief rationale, shown as a non-authoritative hint.
\end{enumerate}

\smallskip
\noindent
\textbf{Full Instructions Provided.}
The following is the full text of the instructions provided for annotation:

\noindent
\emph{[You are provided with] a browsable HTML file with 20 generated languages for the eval. Each language has its description (phonology and grammar) followed by the translations of ten sentences. This time, each sentence translation is shown with the result of an LLM judge which tries to decide if the translation is consistent with the language description -- it judges it as either consistent, mostly consistent, or inconsistent, along with its reasoning. It may sometimes be wrong, but you can use it as a hint to try to decide if a translation is valid.}

\emph{[In the provided spreadsheet], for each sentence in each language, please enter 0 if the sentence is inconsistent with the language, 1 if it is mostly consistent, and 2 if it is fully consistent with the language.}

\emph{Each translation has the following format:}
\emph{
\begin{itemize}
    \item Top, in italics: The English sentence to be translated.
    \item Bold: The attempted translation of this sentence into the language
    \item Typewriter font: The attempted interlinear gloss of this translation.
    \item "New Words": If the language description was missing some words required to translate the sentence, this should list new words that were invented for the language to be able to translate the sentence.
    \item "New Grammar Rules": If the language description was missing some grammar points needed to be able to translate the sentence, this lists these new invented grammar points ("rule"), along with explanations ("justification") for why they should make sense given the language's description.
    \item LLM judgement -- inconsistent / mostly consistent / consistent + explanation. As stated above, these are not guaranteed to be correct, but may help you decide if a translation is really consistent with the language or not.
\end{itemize}
}

\emph{
Important guidelines:
\begin{itemize}
\item Consider a translation valid as long as it is consistent with the language's description, even if new words and/or grammar points had to be invented to translate it.
\item Different translations of different sentences might make different decisions for new words and grammar. They do not need to be consistent with each other -- only judge if a single translation is consistent with the language's original description.
\end{itemize}
}

\input{sec/xx_app_prompts}

\input{sec/xx_app_full_sketch}

%% file: tables/wals_features.tex
\begin{table*}[ht]
\centering
\begin{tabularx}{\textwidth}{|r|l|X|}
\hline
\textbf{WALS\#} & \textbf{Feature} & \textbf{Description/Options} \\
\hline
1 & Consonant inventory & Small ($\leq 20$) vs. Large ($> 20$) \\
\hline
2 & Vowel inventory & Large ($\geq 9$) vs. Small ($< 9$) \\
\hline
13 & Tone & Tonal vs. non-tonal systems \\
\hline
20 & Morphological fusion & Isolating, agglutinating, fusional, etc. \\
\hline
26 & Affixation balance & Relative preference for prefixing vs. suffixing \\
\hline
30 & Gender inventory & Number of gender distinctions (including none) \\
\hline
49 & Case inventory & Minimal, moderate, or extensive case systems \\
\hline
69 & Numeral classifiers & Classifier vs. non-classifier systems \\
\hline
81 & Basic word order & SVO, SOV, VSO, etc. \\
\hline
85 & Adposition type & Prepositions vs. postpositions \\
\hline
86 & Genitive–noun order & Genitive before vs. after noun \\
\hline
87 & Adjective–noun order & Adjectives precede or follow nouns \\
\hline
90 & Relative-clause order & Head-initial vs. head-final \\
\hline
98 & Alignment typology & Nominative–accusative vs. active–stative \\
\hline
107--111 & Valence morphology & Presence/absence of causative, passive, applicative \\
\hline
116 & Question-marking strategy & Interrogative particle vs. inversion \\
\hline
\end{tabularx}
\caption{Linguistic features for typological analysis, with WALS numbers (WALS\#) listed.}
\label{tab:linguistic_features}
\end{table*}

%% file: tables/qual_controlled.tex
\begin{table*}[h!]
\centering
\small

\begin{tabularx}{\linewidth}{@{}>{\raggedright\arraybackslash}X>{\raggedright\arraybackslash}X>{\raggedright\arraybackslash}X@{}}
\toprule
\textbf{User-Input Control} & \textbf{Sample Sentence} & \textbf{Notes} \\
\midrule
The language is a creole combining Japanese and Esperanto. &
\textipa{mat:a-o amai p\~{a}no-un Pamiko}
\newline wait-NPST sweet bread-ACC friend
\newline ``A friend waits for the sweet bread.''
&
Cf. Esperanto \emph{amiko} ``friend'', \emph{pano}, ``bread''; Japanese \emph{matta} ``waited'', \emph{amai} ``sweet'', \emph{pan} ``bread''
\\
\midrule
This language is nasal-centric, with all consonants and vowels nasalized and a rich vocabulary and grammar centered around smells. &
\textipa{\super{n}||\v{@}m-\super{N}g\={@} m\'{a}-\super{n}|@:n-N\`{a}n}
\newline sleep-3SG.I.AGR I.SG-person-NOM
\newline ``The person is sleeping (I can smell their sleeping scent).''
&
Smell-based grammar includes smell-based noun classes and olofactory evidentiality (for events experienced by smell).
\\
\midrule
All inflectional forms are suppletive, and words in a sentence are ordered alphabetically. &
\textipa{janta kasa sake}
\newline see[3S.S/3S.O,INFER] person.ACC person.NOM
\newline ``The person sees the person (I infer).'' & Example sentence shows suppletion (of ``person'') and is in alphabetical order: j < k < s. Other suppletive forms of ``to see'' include \textipa{fans} (see[3S.S/3S.O,VIS]), \textipa{teso} (seeing.NMLZ), \textipa{pote} (see[3S.S/3S.O,REPOR]), and more. \\
\midrule
Verbless constructions (null verb) are common and have a variety of meanings depending on the argument frame present, being used for many functions that are typically lexical verbs (far beyond copular functions).
&
\textipa{st\super{w}\~{u} k\super{j}o.ne m\~{e}.l\super{j}a \|[na.p\super{j}\~{i}.na \!ba p\super{w}jel\super{j}}
\newline PFV.PST ERG.III teacher(III) DAT.III person(III) water(IV)
\newline ``The teacher made the person drink water.''
&
Lit. ``(Did) teacher to person water''; \textipa{st\super{w}\~{u}} is a TAM particle and the predicate ``made drink'' is implied.
\\
\midrule
Triple center embeddings are common. &
\textipa{p\super{h}u:"mor-wo=o [p\super{h}esi-e=je [o"q\super{h}Os-om=jom t\super{h}e"k\super{h}i:sir-na] je"kwe:t\super{h}is-na] ru"p\super{h}uN-om=om o"xo: \t{ts}\super{h}En}
\newline man-AGT=3A [dog-INTR=LOGO.S [rock-PAT=LOGO.P chase-CONV] see-CONV] food-PAT=3P FUT eat
\newline ``The man, who sees his dog which is chasing his (the man's) rock, will eat the food.''
&
From the grammar: ``A defining syntactic feature is its routine use of deep center-embedding... The system allows for multiple, nested embeddings, with triple center-embeddings being common in narrative and formal speech.'' \\
\bottomrule
\end{tabularx}
\caption{\textbf{Qualitative examples} of ConlangCrafter-generated languages with user-input conditions, demonstrating user control and the ability to test creative, linguistic, and philosophical ideas.}
\label{tab:qual_controlled}
\vspace{-1em}
\end{table*}

%% file: sec/xx_app_prompts.tex
\begin{figure*}[!htb]
\small
\centering
\begin{tabular}{|p{2\columnwidth}|}
\hline
\texttt{
\input{prompts/phonology/phon_step1_checklist.txt}
}
\\
\hline
\texttt{
\input{prompts/phonology/phon_step2_summary.txt}
}
\\
\hline
\texttt{
\input{prompts/phonology/phon_step3_word_shapes.txt}
}
\\
\hline
\end{tabular}
\caption{Phonology prompts.}
\label{fig:phonology_prompts}
\end{figure*}

\begin{figure*}[!htb]
\small
\centering
\begin{tabular}{|p{2\columnwidth}|}
\hline
\texttt{
\input{prompts/grammar/gram_step1_checklist.txt}
}
\\
\hline
\texttt{
\input{prompts/grammar/gram_step2_summary.txt}
}
\\
\hline
\texttt{
\input{prompts/grammar/gram_step3_expand.txt}
}
\\
\hline
\texttt{
\input{prompts/grammar/merge_sections.txt}
}
\\
\hline
\end{tabular}
\caption{Grammar prompts.}
\label{fig:grammar_prompts}
\end{figure*}

\begin{figure*}[!htb]
\small
\centering
\begin{tabular}{|p{2\columnwidth}|}
\hline
\texttt{
\input{prompts/lexicon/lex_step1_extract.txt}
}
\\
\hline
\texttt{
\input{prompts/lexicon/lex_step2_expand.txt}
}
\\
\hline
\end{tabular}
\caption{Lexicon prompts.}
\label{fig:lexicon_prompts}
\end{figure*}

\begin{figure*}[!htb]
\small
\centering
\begin{tabular}{|p{2\columnwidth}|}
\hline
\texttt{
\input{prompts/qa/qa_critic.txt}
}
\\
\hline
\texttt{
\input{prompts/qa/qa_critic_with_context.txt}
}
\\
\hline
\texttt{
\input{prompts/qa/qa_amend.txt}
}
\\
\hline
\end{tabular}
\caption{Self-refinement prompts.}
\label{fig:sr_prompts}
\end{figure*}

\begin{figure*}[!htb]
\small
\centering
\begin{tabular}{|p{2\columnwidth}|}
\hline
\texttt{
\input{prompts/qa/qa_translation_critic_with_context.txt}
}
\\
\hline
\texttt{
\input{prompts/qa/qa_translation_amend.txt}
}
\\
\hline
\end{tabular}
\caption{Self-refinement prompts (translation).}
\label{fig:sr_translation_prompts}
\end{figure*}

\begin{figure*}[!htb]
\small
\centering
\begin{tabular}{|p{2\columnwidth}|}
\hline
\texttt{
\input{prompts/translation/translation_single.txt}
}
\\
\hline
\end{tabular}
\caption{Translation prompt.}
\label{fig:translation_prompt}
\end{figure*}

\begin{figure*}[!htb]
\small
\centering
\begin{tabular}{|p{2\columnwidth}|}
\hline
\texttt{
\input{prompts/full_pipeline.txt}
}
\\
\hline
\end{tabular}
\caption{Baseline single prompt.}
\label{fig:baseline_prompt}
\end{figure*}

\begin{figure*}[!htb]
\small
\centering
\begin{tabular}{|p{2\columnwidth}|}
\hline
{\tiny
\texttt{
\input{prompts/evaluation/diversity_medium.txt}
}
}
\\
\hline
\end{tabular}
\caption{Diversity evaluation prompt.}
\label{fig:diversity_prompt}
\end{figure*}

\begin{figure*}[!htb]
\small
\centering
\begin{tabular}{|p{2\columnwidth}|}
\hline
\texttt{
\input{prompts/evaluation/translation_consistency.txt}
}
\\
\hline
\end{tabular}
\caption{Consistency evaluation prompt.}
\label{fig:consistency_prompt}
\end{figure*}

%% file: prompts/phonology/phon_step1_checklist.txt
I am designing a hypothetical language’s phonology. Provide me with \{n\_questions\} sliders with scales from 1 to \{scale\_size\}, and/or multiple-choice questions with \{n\_answers\} possible answers (numbered 1-\{scale\_size\}), to determine the most important typological features to construct this language's phonology. There should be \{n\_questions\} of them total (counting sliders and multiple-choice questions together).

Provide no additional explanation or discussion.

%% file: prompts/phonology/phon_step2_summary.txt
I am designing a hypothetical language’s phonology.

Consider the following checklist:

== START CHECKLIST ==

\{checklist\}

== END CHECKLIST ==

Use values \{values\} for those respectively.

It should also obey the following constraint(s); if these contradict the above, these take priority:

\{custom\}

Now write a summary of its phonology.

Make sure there are a couple of aspects that are unusual, creative, interesting, or surprising, while still obeying the constraints from above.

Format your output as follows. Give no additional explanation or discussion.

\# Phonology

\#\# Consonants

(IPA chart of consonants, as markdown table)

\#\# Vowels

(IPA chart of vowels, as markdown table)

\#\# Phonotactics

(Brief, single-paragraph explanation)

\#\# Suprasegmentals

(Brief, single-paragraph explanation)

%% file: prompts/phonology/phon_step3_word_shapes.txt
A hypothetical language has the following phonology:

=== START ===

\{phonology\}

=== END ===

Write a description of the distribution of word shapes in this language, touching content vs. function items and the distribution of word lengths and syllable counts. Use qualitative descriptors (like "most," "many," "some," "few," "rare") rather than specific percentages or numbers.

Include at least \{n\} diverse lexical items that illustrate the points discussed.

Give the words in IPA (underlying phonemic representation only), making sure all phonemic features are indicated, including contrastive suprasegmentals (if relevant). Do not give their translations in English; only note if they are content or function items and how common or uncommon they are.

The new lexical items should obey the following constraints:

\{custom\}

If these contradict anything above, these constraints take priority.

Format your output as follows. Give no additional explanation or discussion.

\#\# Word Shapes and Lexical Statistics

(description here)

%% file: prompts/grammar/gram_step1_checklist.txt
I am designing a hypothetical language's grammar (morphology and syntax). Provide me with \{n\_questions\} sliders with scales from 1 to \{scale\_size\}, and/or multiple-choice questions with \{n\_answers\} possible answers (numbered 1-\{scale\_size\}), to determine the most important typological features to construct this language's grammar (morphology and syntax). There should be \{n\_questions\} of them total (counting sliders and multiple-choice questions together).

Provide no additional explanation or discussion.

%% file: prompts/grammar/gram_step2_summary.txt
I am designing a hypothetical language’s grammar (morphology and syntax).

Consider the following checklist:

== START CHECKLIST ==

\{checklist\}

== END CHECKLIST ==

Use values \{values\} for those respectively.

It should also obey the following constraint(s); if these contradict the above, these take priority:

\{custom\}

The language has the following phonology:

=== START ===

\{phonology\}

=== END ===

Now write a summary of its grammar (morphology and syntax).

Make sure there are a couple of aspects that are unusual, creative, interesting, or surprising, while still obeying the constraints from above.

For any statement, include examples with interlinear glosses and English translations. When possible, use words/roots from the list above in examples. Examples should be provided right next to the statements that they illustrate (not all at the end).

Format your output as follows. Give no additional explanation or discussion.

\# Grammar

\#\# Morphology

(summary of morphology here)

\#\# Syntax

(summary of syntax here)

%% file: prompts/grammar/gram_step3_expand.txt
A hypothetical language has the following phonology:

=== START ===

\{phonology\}

=== END ===

Here is a summary of its grammar (morphology and syntax):

=== START ===

\{grammar\}

=== END ===

Write expanded grammar sections including important points that are missing above, that would be needed to understand or use the language.

Make sure there are a couple of aspects that are unusual, creative, interesting, or surprising, while still obeying the constraints from above.

It should also obey the following constraint(s); if these contradict the above, these take priority:

\{custom\}

For any statement, include examples with interlinear glosses and English translations. When possible, use words/roots from the list above in examples. Examples should be provided right next to the statements that they illustrate (not all at the end).

%% file: prompts/grammar/merge_sections.txt
Combine the summaries below together. If there are any contradictions, amend as needed to make everything consistent. Make sure to include all information and examples that appear anywhere in any of the summaries.

Only return the new summary (do not give any further commentary or explain your amendations).

\{summaries\}

Format your output as follows. Make sure it contains all information from the summaries above. Examples include both interlinear glosses and English translations, and should be provided right next to the statements that they illustrate (not all at the end). Give no additional explanation or discussion.

\# Grammar

\#\# Morphology

(morphological information here)

\#\# Syntax

(syntactic information here)

%% file: prompts/lexicon/lex_step1_extract.txt
A hypothetical language has the following phonology and grammar:

=== START ===

\{phonology\}

\{grammar\}

=== END ===

Output a CSV with a lexicon of the language, including all lexical items that appear in the language's information above.

The CSV column names and their contents should be:

* "ipa": Lexical item as IPA (without brackets)

* "pos": Part of speech

* "translation": Translation of item into English

* "grammar": Any grammar points or information specific to that lexical item (e.g. inflectional or conjugational classes, irregularities, ...)

* "derivation": If the item is derived, note its derivation here

* "notes": Any additional notes

Provide translations for all lexical items, even those whose translations are not explicitly specified in the provided background above (i.e. don't list any translations as "unknown").

Do not provide any further explanation or text, only the CSV.

%% file: prompts/lexicon/lex_step2_expand.txt
A hypothetical language has the following phonology:

=== START ===

\{phonology\}

=== END ===

It has the following grammar:

=== START ===

\{grammar\}

=== END ===

It has the following lexicon:

=== START ===

\{lexicon\}

=== END ===

Propose at least \{n\} additional lexical items to add to the lexicon, including a diverse set of items covering common and uncommon concepts and different parts of speech. Make sure they obey the language's phonology and morphology, including following the word shape distribution described in "Word Shapes and Lexical Statistics". Also make sure they do not overlap any existing lexical items.

Output a CSV in the same format as the lexicon above. The CSV column names and their contents should be:

* "ipa": Lexical item as IPA (without brackets)

* "pos": Part of speech

* "translation": Translation of item into English

* "grammar": Any grammar points or information specific to that lexical item (e.g. inflectional or conjugational classes, irregularities, ...)

* "derivation": If the item is derived, note its derivation here

* "notes": Any additional notes

Do not provide any further explanation or text, only the CSV.

%% file: prompts/qa/qa_critic.txt
Here is a hypothetical language's \{content\_type\}:

=== START ===

\{content\}

=== END ===

Is the description of the language's \{content\_type\} consistent? Return a JSON with keys:

* "overall\_score": Integer from 1 (completely inconsistent) to 10 (completely consistent), scoring the \{content\_type\} overall. If there are not issues that are strictly contradictions or errors (for example, if the only issues are ambiguities/unclear points), the score should be 9 (consistent but somewhat unclear). If there are only minor contradictions or errors, the score should be 8 (very minor inconsistencies).

* "issues": List of errors and inconsistencies found (if any) and their corrections. Each item in this list should be an object with the following keys and values:

  * "issue": String describing issue

  * "type": One of "inconsistency"/"error"/"ambiguity", indicating if the issue is an inconsistency (contradiction), error (other mistake), or ambiguity (not strictly a mistake but unclear)

  * "correction": String describing how it should be corrected

  * "priority": 1 (high -- severe issue), 2 (medium), or 3 (low -- minor or very minor issue)

Be as exhaustive as possible in your search for issues.

Output only this JSON object. Give no additional explanation or discussion.

%% file: prompts/qa/qa_critic_with_context.txt
Here is a hypothetical language's \{content\_type\}:

=== START ===

\{content\}

=== END ===

For context, here is the language's \{context\_type\}:

=== START ===

\{context\}

=== END ===

Is the description of the language's \{content\_type\} consistent with itself and with its \{context\_type\}? Return a JSON with keys:

* "overall\_score": Integer from 1 (completely inconsistent) to 10 (completely consistent), scoring the \{content\_type\} overall. If there are not issues that are strictly contradictions or errors (for example, if the only issues are ambiguities/unclear points), the score should be 9 (consistent but somewhat unclear). If there are only minor contradictions or errors, the score should be 8 (very minor inconsistencies).

* "issues": List of errors and inconsistencies found (if any) and their corrections. Each item in this list should be an object with the following keys and values:

  * "issue": String describing issue

  * "type": One of "inconsistency"/"error"/"ambiguity", indicating if the issue is an inconsistency (contradiction), error (other mistake), or ambiguity (not strictly a mistake but unclear)

  * "correction": String describing how it should be corrected in the \{content\_type\}. Note that only the \{content\_type\} can be corrected, not the \{context\_type\}. If the issue is due to a conflict with the \{context\_type\}, make sure this description will be clear to an editor who does not have the \{context\_type\} in front of them when amending the \{content\_type\}.

  * "priority": 1 (high -- severe issue), 2 (medium), or 3 (low -- minor or very minor issue)

Be as exhaustive as possible in your search for issues.

Output only this JSON object. Give no additional explanation or discussion.

%% file: prompts/qa/qa_amend.txt
Here is a hypothetical language's phonology:

=== START ===

\{content\}

=== END ===

Here is a judgement of its overall consistency on a scale of 1 (completely inconsistent) - 5 (completely consistent), and specific issues found (if any) and their priorities (from 1=high to 3=low) and how to correct them:

\{judgement\}

Correct these points (and any other errors or inconsistencies, if any), outputting an amended version (without === START === / === END === lines). Give no additional explanation or discussion.

%% file: prompts/qa/qa_translation_critic_with_context.txt
Here is a translation result for a constructed language:

=== START ===

\{content\}

=== END ===

For context, here is the language specification:

=== START ===

\{context\}

=== END ===

Evaluate the quality and accuracy of this translation. Return a JSON with keys:

* "overall\_score": Integer from 1 (completely incorrect) to 10 (excellent translation), scoring the translation overall. Consider:

  - Adherence to phonological rules (10: fully consistent, 8-9: minor violations, 5-7: some violations, 1-4: major violations)

  - Adherence to grammatical rules (10: fully consistent, 8-9: minor violations, 5-7: some violations, 1-4: major violations)

  - Appropriate use of lexicon (10: excellent use, 8-9: mostly appropriate, 5-7: some issues, 1-4: poor lexical choices)

  - Quality of glossing (10: accurate and complete, 8-9: mostly accurate, 5-7: some errors, 1-4: poor glossing)

  - Semantic accuracy (10: meaning preserved, 8-9: mostly preserved, 5-7: some meaning lost, 1-4: meaning significantly altered)

* "issues": List of problems found (if any). Each item should be an object with:

  * "issue": String describing the problem

  * "type": One of "phonological\_violation"/"grammatical\_violation"/"lexical\_error"/"glossing\_error"/
  
  "semantic\_error"/"consistency\_error"

  * "correction": String describing how the translation should be corrected

  * "priority": 1 (high -- major error affecting meaning or violating core rules), 2 (medium -- noticeable error), or 3 (low -- minor issue or style preference)

Be thorough in checking:

1. Phonological consistency: Do invented words follow the sound patterns and constraints?

2. Grammatical consistency: Are word order, morphology, and syntax rules followed?

3. Lexical appropriateness: Are existing words used correctly? Are new words justified and well-formed?

4. Glossing accuracy: Does the gloss correctly break down morphemes and grammatical functions?

5. Semantic preservation: Does the translation convey the intended meaning?

6. Internal consistency: Are new rules and words consistent with each other?

Output only this JSON object. Give no additional explanation or discussion.

%% file: prompts/qa/qa_translation_amend.txt
Here is a translation result for a constructed language that might need improvement:

=== START ===

\{content\}

=== END ===

Based on this quality assessment:

=== START ===

\{judgement\}

=== END ===

Please provide an improved version of the translation that addresses the identified issues. Make sure to:

1. Fix any phonological violations by adjusting words to follow the sound patterns

2. Correct grammatical errors to match the specified rules

3. Improve lexical choices and justify any new words created

4. Ensure accurate glossing with proper morpheme breakdown

5. Preserve semantic meaning while fixing technical issues

6. Maintain consistency across all elements

Return the corrected translation in the same JSON format as the original:

\{\{

    "sentences": [

        \{\{

            "conlang\_sentence": "<corrected sentence in the constructed language only - no glosses, morpheme breaks, or explanations>",

            "gloss": "<corrected word-by-word breakdown with morphemes and grammatical functions>",

            "new\_words": [\{\{"<new\_word>": "<english\_translation>"\}\}],

            "new\_grammar\_rules": [\{\{"rule": "<description\_of\_new\_rule>", "justification": "<explanation\_of\_why\_needed\_and\_consistency>"\}\}]

        \}\}

    ]

\}\}

Output only this JSON object. Give no additional explanation or discussion.

%% file: prompts/translation/translation_single.txt
You are a skilled linguist working with a constructed language. You have been provided with a complete language specification including vocabulary, grammar and phonology.

A hypothetical language has the following phonology:

=== START ===

\{phonology\}

=== END ===

It has the following grammar:

=== START ===

\{grammar\}

=== END ===

\{lexicon\_section\}

Your task is to translate the following English sentence into this constructed language:

English sentence: \{input\_sentence\}

Instructions:

1. Use the vocabulary provided in the language specification (if available)

2. Follow the grammatical rules and patterns described

3. Apply the phonological conventions

4. If a word is not in the vocabulary, invent a new word that is consistent with the specifications

5. If a grammatical construction is needed but not specified, invent a new grammar rule that is consistent with the existing language patterns

6. Ensure the translation follows the word order and morphological patterns specified

If you create new words, ensure they:

- Follow the phonological constraints of the language

- Have appropriate semantic scope for the meaning needed

- Use appropriate derivational processes if applicable

If you create new grammar rules, ensure they:

- Are consistent with existing grammatical patterns

- Follow the language's morphological and syntactic tendencies

- Are plausible extensions of the existing system

Provide a gloss line: break down the conlang sentence word by word, showing morphemes and their grammatical functions using standard linguistic abbreviations (e.g., FUT, SBJ, OBJ, ACC, etc.). The gloss should use English abbreviations and morpheme breakdowns, NOT the conlang words.

Return the result in JSON format only:

\{\{

    "sentences": [

        \{\{

            "conlang\_sentence": "<sentence in the constructed language>",

            "gloss": "<word-by-word breakdown with morphemes and grammatical functions>",

            "new\_words": [\{\{"<new\_word>": "<english\_translation>"\}\}],

            "new\_grammar\_rules": [\{\{"rule": "<description\_of\_new\_rule>", "justification": "<explanation\_of\_why\_needed\_and\_consistency>"\}\}]

        \}\}

    ]

\}\}

%% file: prompts/full_pipeline.txt
Create a conlang (constructed language). Include a description of its phonology, grammar (morphology and syntax), and lexicon. Use the following format:

\# Phonology

\#\# Consonants

(IPA chart of consonants, as markdown table)

\#\# Vowels

(IPA chart of vowels, as markdown table)

\#\# Phonotactics

(Brief, single-paragraph explanation)

\#\# Suprasegmentals

(Brief, single-paragraph explanation)

\#\# Word Shapes and Lexical Statistics

(description here)

\# Grammar

\#\# Morphology

(morphological information here in markdown format)

\#\# Syntax

(syntactic information here)

\# Lexicon

(list of at least 100 lexical items in the language in a csv format)

\# Corpus

give translations and interlinear glosses for the following 10 sentences. Invent new words or grammar rules, if needed, for each sentence.

Provide a gloss line: break down the conlang sentence word by word, showing morphemes and their grammatical functions using standard linguistic abbreviations (e.g., FUT, SBJ, OBJ, ACC, etc.). The gloss should use English abbreviations and morpheme breakdowns, NOT the conlang words.

1. The big dog is sleeping.

2. Where is my book?

3. She will give him water.

4. This child walked to that house yesterday

5. Are you hungry?

6. Give me the red bird!

7. The woman and the man are talking.

8. I do not see three cats.

9. There is a black mountain.

10. The two children played in the garden.

Return the translation results in JSON format:

\{\{

    "sentences": [

        \{\{

            "english\_sentence": "<original English sentence 1>",

            "conlang\_sentence": "<sentence in the constructed language>",

            "gloss": "<word-by-word breakdown with morphemes and grammatical functions>",

            "new\_words": [\{\{"<new\_word>": "<english\_translation>"\}\}],

            "new\_grammar\_rules": [\{\{"rule": "<description\_of\_new\_rule>", "justification": "<explanation\_of\_why\_needed\_and\_consistency>"\}\}]

        \}\},

        \{\{

            "english\_sentence": "<original English sentence 2>",

            "conlang\_sentence": "<sentence in the constructed language>",

            "gloss": "<word-by-word breakdown with morphemes and grammatical functions>",

            "new\_words": [\{\{"<new\_word>": "<english\_translation>"\}\}],

            "new\_grammar\_rules": [\{\{"rule": "<description\_of\_new\_rule>", "justification": "<explanation\_of\_why\_needed\_and\_consistency>"\}\}]

        \}\}

    ]

\}\}

%% file: prompts/evaluation/diversity_medium.txt
Please analyze the following documentation for the constructed language "\{your\_language\_name\}" and return the typological analysis as a JSON object.

A hypothetical language has the following phonology:

=== START ===

\{phonology\}

=== END ===

It has the following grammar:

=== START ===

\{grammar\}

=== END ===

It has the following lexicon:

=== START ===

\{lexicon\}

=== END ===

\#\# JSON Schema to Populate

\{

  "language\_name": "\{your\_language\_name\}",

  "analysis\_timestamp\_utc": "\{current\_utc\_iso\_timestamp\}",

  "typology": \{

    "wals\_81A\_svo\_order": \{

      "value": "SVO | SOV | VSO | VOS | OVS | OSV | No Dominant Order | null",

      "confidence": "High | Medium | Low"

    \},

    "wals\_85A\_adpositions": \{

      "value": "Prepositions | Postpositions | Inpositions | No Adpositions | null",

      "confidence": "High | Medium | Low"

    \},

    "wals\_87A\_adjective\_noun": \{

      "value": "Adjective-Noun | Noun-Adjective | null",

      "confidence": "High | Medium | Low"

    \},

    "wals\_20A\_fusion": \{

      "value": "Isolating | Agglutinative | Fusional | null",

      "confidence": "High | Medium | Low"

    \},

    "wals\_26A\_affixation": \{

      "value": "Strongly Suffixing | Weakly Suffixing | Equal Prefixing/Suffixing | Weakly Prefixing | Strongly Prefixing | null",

      "confidence": "High | Medium | Low"

    \},

    "wals\_13A\_tone": \{

      "value": "Tonal | Non-tonal | null",

      "confidence": "High | Medium | Low"

    \},

    "wals\_30A\_gender\_count": \{

      "value": "None | Two | Three | Four | Five or more | null",

      "confidence": "High | Medium | Low"

    \},

    "wals\_49A\_case\_count": \{

      "value": "None (0-1) | Minimal (2-3) | Moderate (4-5) | Extensive (6+) | null",

      "confidence": "High | Medium | Low"

    \},

    "wals\_alignment": \{

      "value": "Accusative | Ergative | Active-Stative | Split | null",

      "confidence": "High | Medium | Low"

    \},

    "wals\_87B\_relative\_clause\_order": \{

      "value": "Head-initial | Head-final | Mixed | null",

      "confidence": "High | Medium | Low"

    \},

    "wals\_genitive\_order": \{

      "value": "Genitive-Noun | Noun-Genitive | null",

      "confidence": "High | Medium | Low"

    \},

    "wals\_question\_marking": \{

      "value": "Intonation | Particle | Morphological | Inversion | null",

      "confidence": "High | Medium | Low"

    \},

    "phoneme\_inventory": \{

      "vowels": "Small ($\leq$5) | Medium (6–8) | Large ($\geq$9) | null",

      "consonants": "Small ($\leq$20) | Medium (21–35) | Large ($\geq$36) | null",

      "confidence": "High | Medium | Low"

    \},

    "valence\_morphology": \{

      "causative": "Yes | No | null",

      "passive": "Yes | No | null",

      "applicative": "Yes | No | null"

    \},

    "numeral\_classifiers": \{

      "value": "Classifier language | Non-classifier | null",

      "confidence": "High | Medium | Low"

    \}

  \}

\}

%% file: prompts/evaluation/translation_consistency.txt
You are a senior linguist specializing in constructed languages. Your task is to check the internal consistency of a single translated sentence from a conlang, examining how well the translation aligns with the given phonology, grammar, and lexicon.

1. Carefully read the three language sections (PHONOLOGY, GRAMMAR, LEXICON).

2. Examine the provided sentence translation, including the conlang sentence, gloss, and claimed new words/grammar rules.

3. Identify every inconsistency, error, or mismatch you can find in this specific translation (for example: phonotactic violations, grammatical rule violations, lexicon mismatches, gloss errors, etc.).

   Only account for linguistic errors, not formatting or stylistic errors.

4. For each issue, decide how serious it is:

   • minor – cosmetic problems that don't affect comprehension or linguistic accuracy

   • moderate – noticeable errors that affect clarity or violate established rules but don't break the translation entirely

   • major – serious violations that make the translation incorrect or incomprehensible given the language system

5. Produce your answer ONLY in the JSON format specified below. No additional keys, no comments, no markdown fences.

Required output JSON format:

\{

  "inconsistencies": [

    \{

      "area": "string (Phonology | Grammar | Lexicon | Translation | Gloss | Cross-domain)",

      "description": "string (describe what is inconsistent)",

      "severity": "string (minor | moderate | major)"

    \}

    … (0 or more items)

  ],

  "final\_verdict": "string (consistent | mostly consistent | inconsistent)"

\}

Guidelines for the final verdict:

• If no inconsistencies are found, return "consistent".

• If only minor issues are found, or one moderate issue, return "mostly consistent".

• If any major issue, or multiple moderate issues are found, return "inconsistent".

===  LANGUAGE DATA STARTS  ===

Phonology:

=== START ===

\{phonology\}

=== END ===

Grammar:

=== START ===

\{grammar\}

=== END ===

Lexicon:

=== START ===

\{lexicon\}

=== END ===

===  TRANSLATION TO EVALUATE  ===

English sentence: \{english\_sentence\}

Conlang translation: \{conlang\_sentence\}

Gloss: \{gloss\}

New words claimed: \{new\_words\}

New grammar rules claimed: \{new\_grammar\_rules\}

%% file: sec/xx_app_full_sketch.tex
\begin{figure*}[p]
    \centering
    \includegraphics[width=\textwidth, page=1]{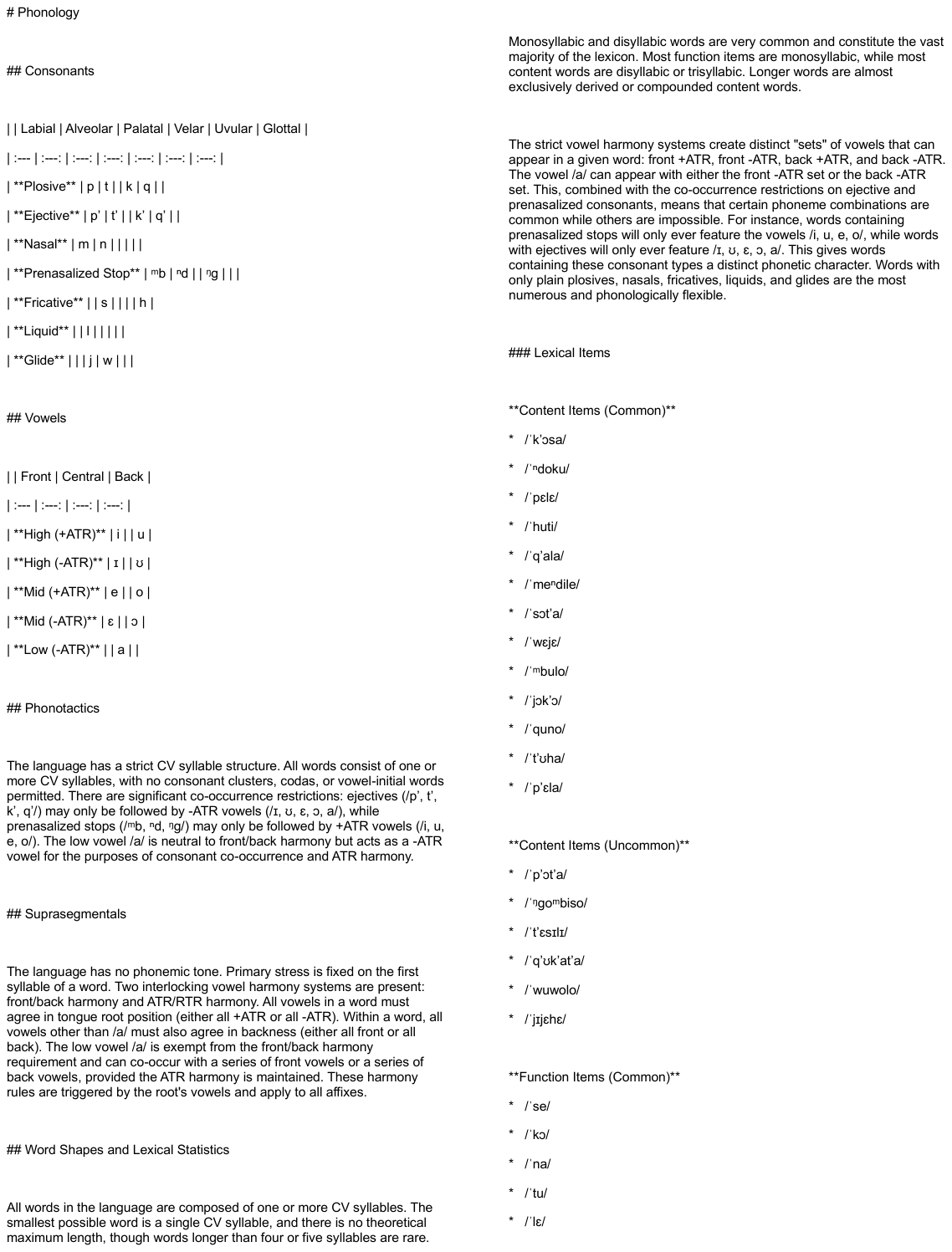}
    \caption{Sample language sketch, page 1/7.}
    \label{fig:sample_sketch_p1}
\end{figure*}

\begin{figure*}[p]
    \centering
    \includegraphics[width=\textwidth, page=2]{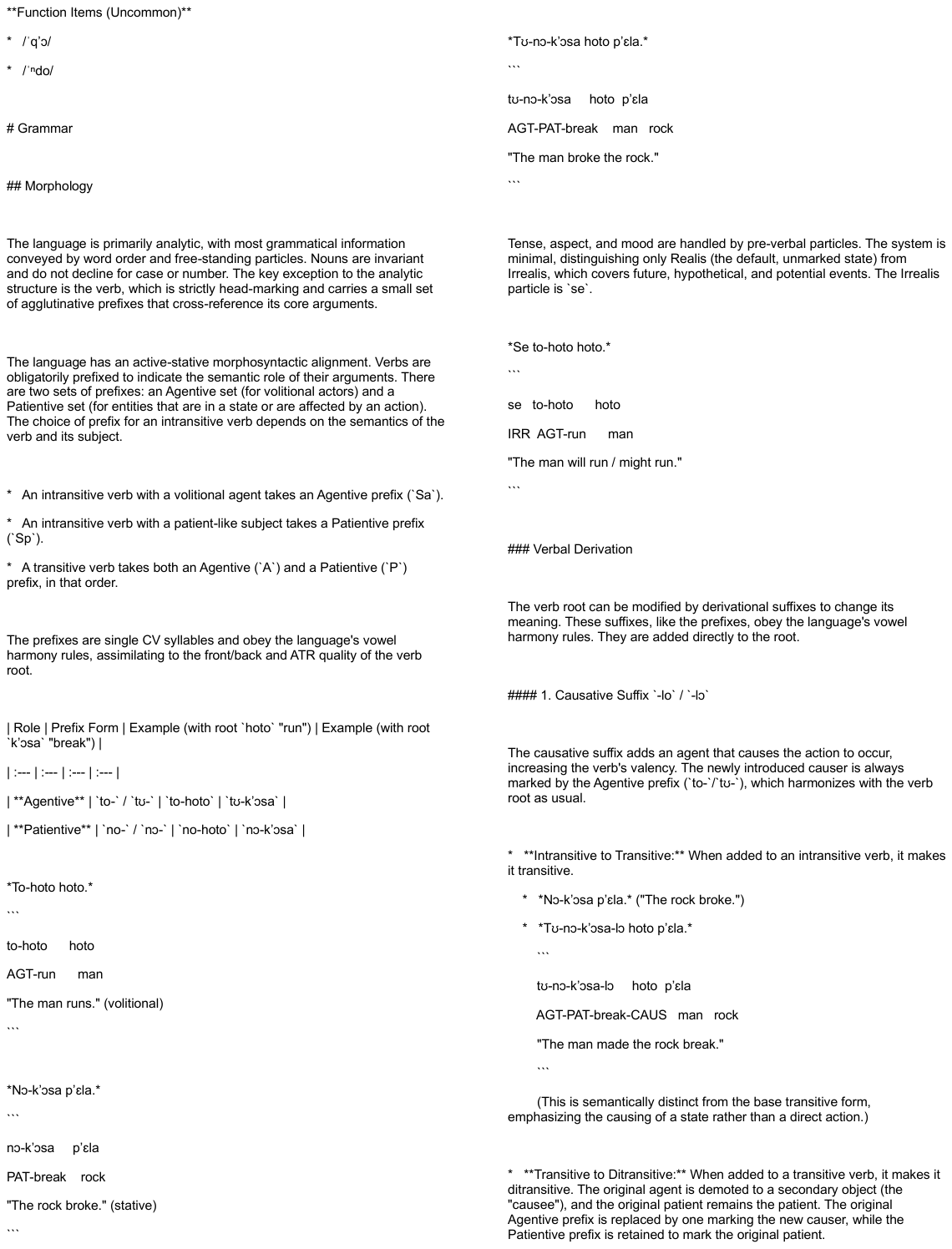}
    \caption{Sample language sketch, page 2/7.}
    \label{fig:sample_sketch_p2}
\end{figure*}

\begin{figure*}[p]
    \centering
    \includegraphics[width=\textwidth, page=3]{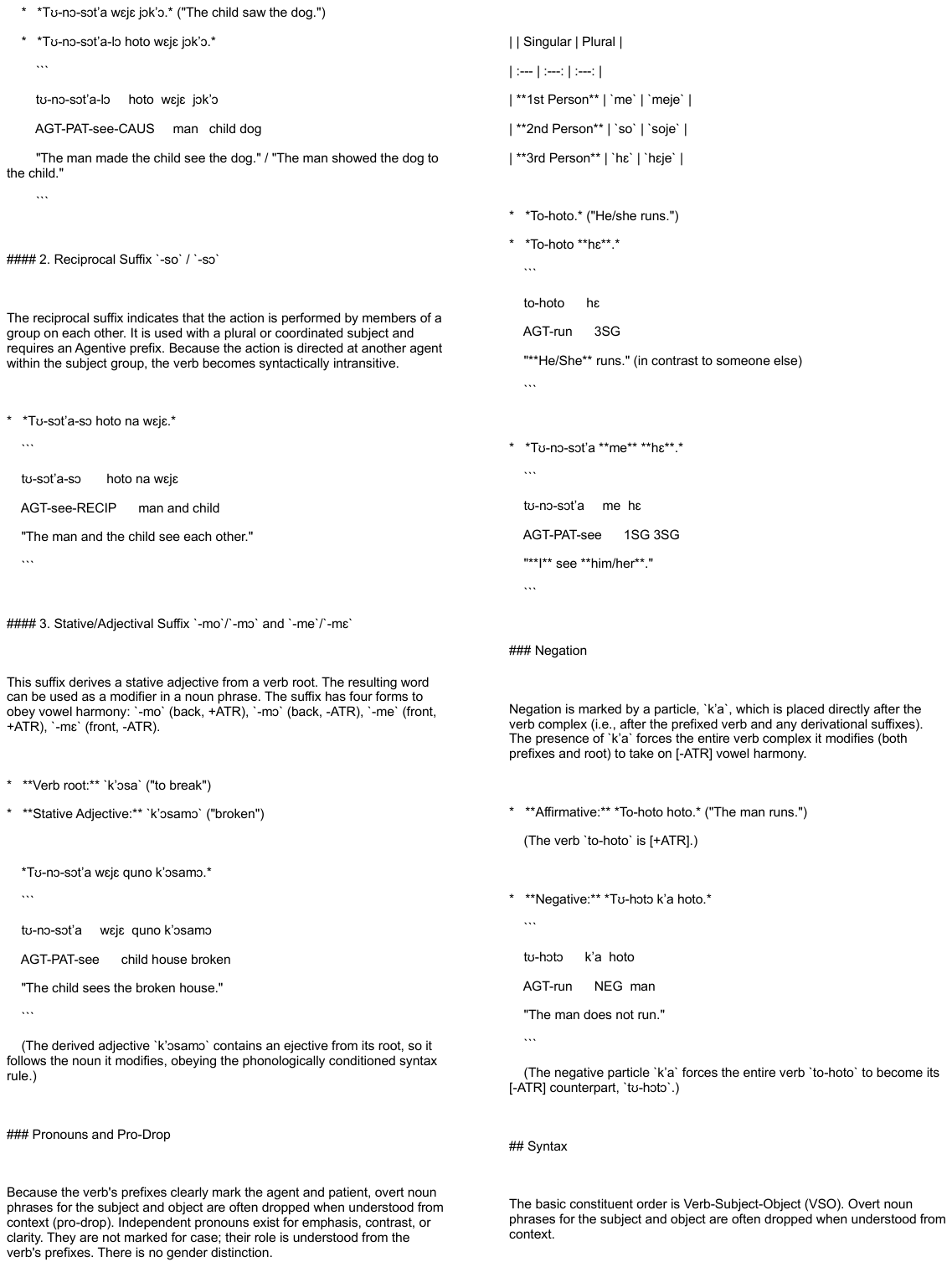}
    \caption{Sample language sketch, page 3/7.}
    \label{fig:sample_sketch_p3}
\end{figure*}

\begin{figure*}[p]
    \centering
    \includegraphics[width=\textwidth, page=4]{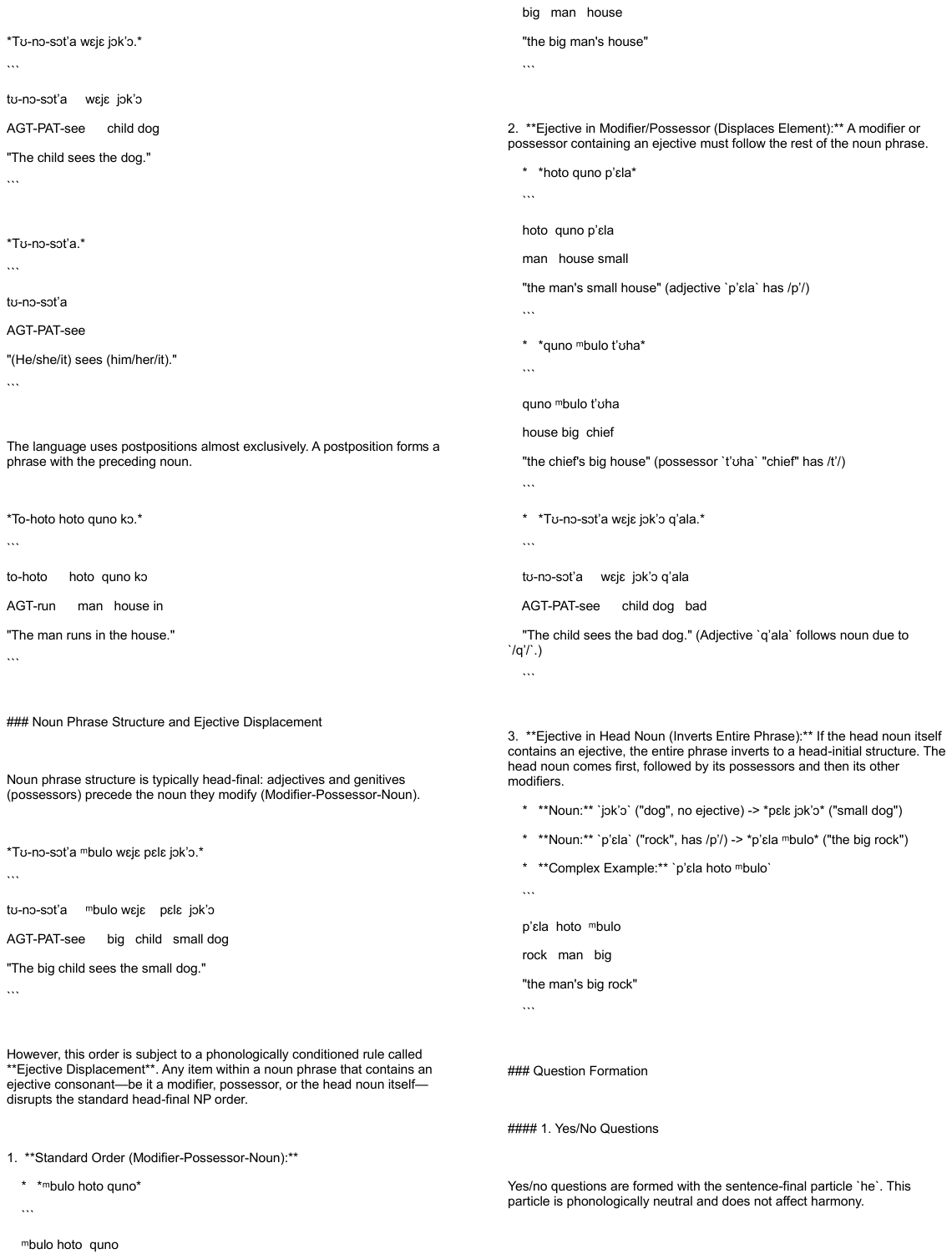}
    \caption{Sample language sketch, page 4/7.}
    \label{fig:sample_sketch_p4}
\end{figure*}

\begin{figure*}[p]
    \centering
    \includegraphics[width=\textwidth, page=5]{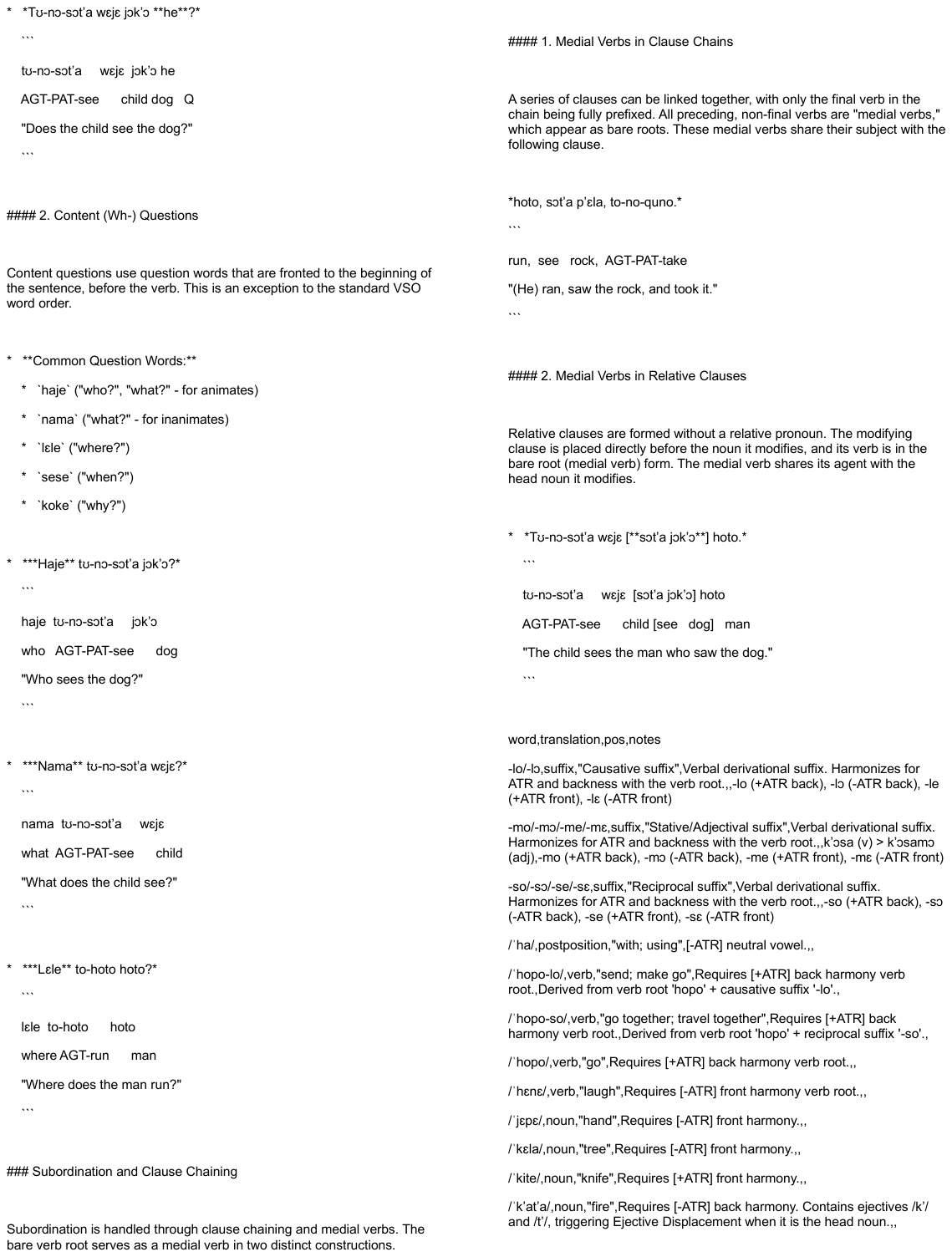}
    \caption{Sample language sketch, page 5/7.}
    \label{fig:sample_sketch_p5}
\end{figure*}

\begin{figure*}[p]
    \centering
    \includegraphics[width=\textwidth, page=6]{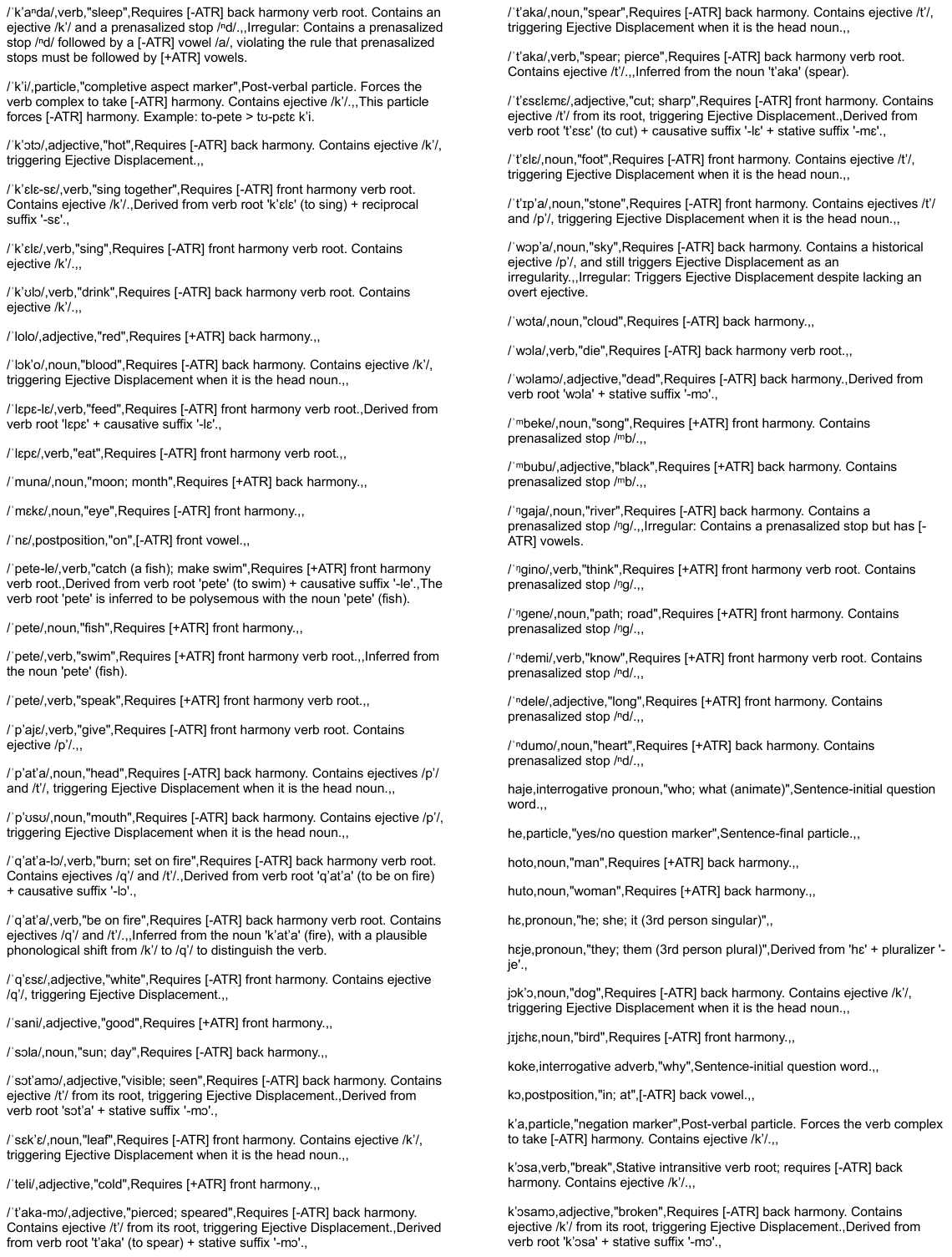}
    \caption{Sample language sketch, page 6/7.}
    \label{fig:sample_sketch_p6}
\end{figure*}

\begin{figure*}[p]
    \centering
    \includegraphics[width=\textwidth, page=7]{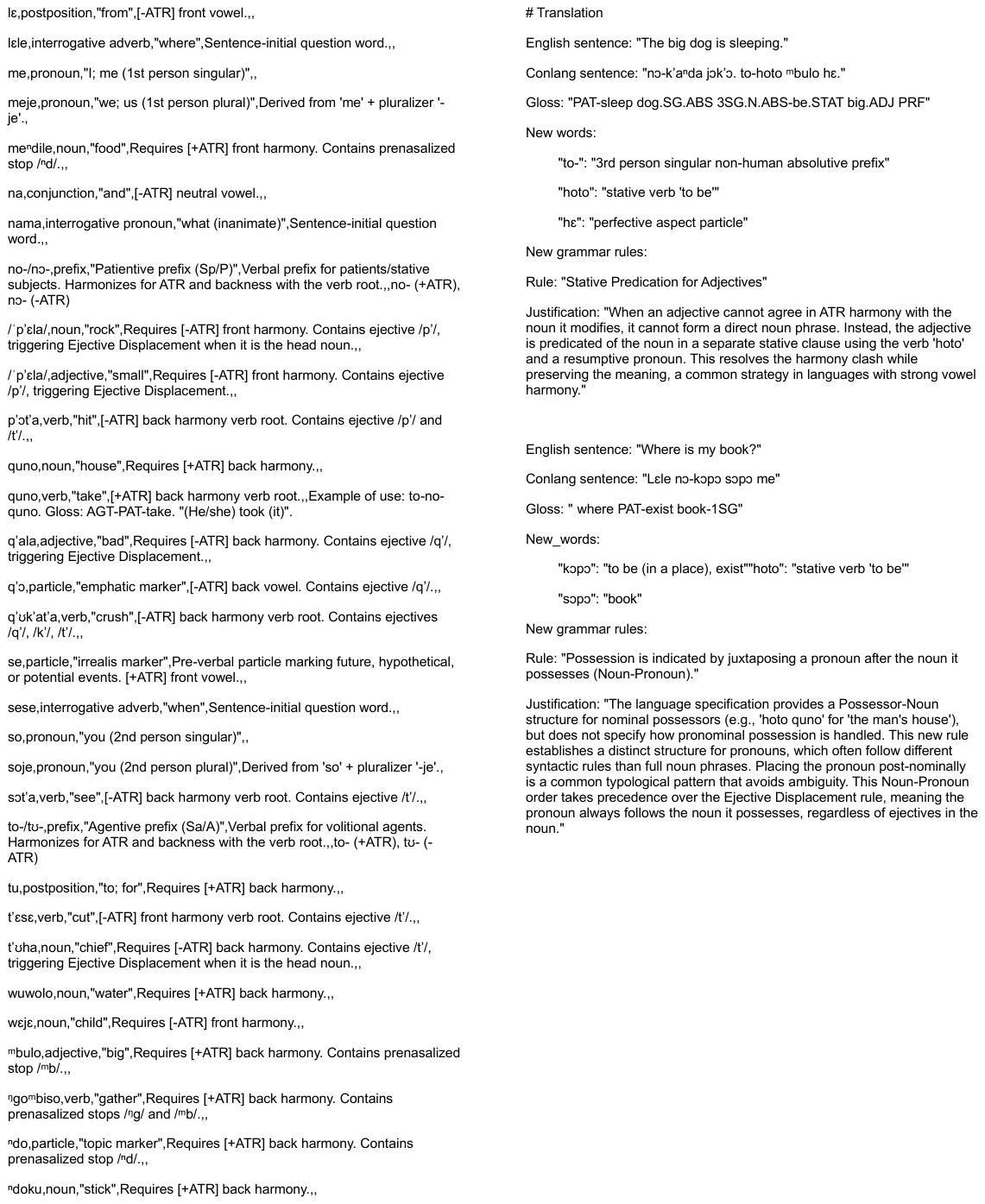}
    \caption{Sample language sketch, page 7/7.}
    \label{fig:sample_sketch_p7}
\end{figure*}